\documentclass{article}
\usepackage[eandd, final]{neurips_2026}  
\usepackage[utf8]{inputenc}
\usepackage[T1]{fontenc}
\usepackage{hyperref}
\usepackage{url}
\usepackage{booktabs}
\usepackage{amsfonts}
\usepackage{amsmath}
\usepackage{amssymb}
\usepackage{nicefrac}
\usepackage{microtype}
\usepackage{xcolor}
\usepackage{graphicx}
\usepackage{svg}
\usepackage{enumitem}
\usepackage{multirow}
\usepackage{threeparttable}

\usepackage{tabularx}
\newcolumntype{Y}{>{\centering\arraybackslash}X}
\usepackage{fontawesome5}  
\newcommand{\hficon}{\raisebox{-0.25ex}{\includegraphics[height=1.1em]{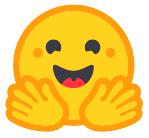}}}
\title{InformedXRD: Reproducible Benchmarks and Physics-Informed Evaluation for Powder Diffraction Symmetry Classification}
\author{%
\begin{minipage}{\textwidth}
\centering
  Edward G. Friedman$^{1,2}$ \quad Elizabeth J. Baggett$^{3,2}$ \quad
  Abhishek Shetty$^{4}$ \quad Derrick Chan-Sew$^{4}$ \\
  Vanellsa Acha$^{4}$ \quad Harshitha Dwaracherla$^{4}$ \quad
  Paul Kienzle$^{2}$ \quad William Ratcliff$^{2,5,6}$\thanks{Corresponding author: \texttt{william.ratcliff@nist.gov}} \\[0.3em]
  $^{1}$School of Computer Science, Carnegie Mellon University;
  $^{2}$NIST Center for Neutron Research;
  $^{3}$Dept.\ of Physics, Boston College;
  $^{4}$School of Information, UC Berkeley;
  $^{5}$Dept.\ of Materials Science and Engineering and $^{6}$Dept.\ of Physics, University of Maryland
\end{minipage}
}
\begin{document}
\maketitle


\begin{abstract}
 
Machine learning for powder diffraction (PXRD) symmetry classification is being trained at scale, yet the field lacks a common evaluation standard: papers report results on different subsets of experimental archives, with different target taxonomies, preprocessing steps, and metrics. Published models are rarely released in a form that permits rescoring. We present \emph{InformedXRD}, an evaluation framework with three parts: (i) a \textbf{machine-readable mapping} from 230 space groups to 99 extinction groups, the finest classes that powder diffraction can identify from systematic absences alone; (ii) a \textbf{standardized benchmark-construction method} whose inclusion rules are specified as code and reproduce from a frozen snapshot, instantiated as RRUFF-473, RRUFF-325, and, as a small example, opXRD-98; (iii) an \textbf{evaluation reporting protocol} including prior-only frequency baselines, a label-permutation test, a hierarchy-aware error metric on the condensed \textit{translationengleiche} subgroup graph, and stratified evaluation. We build on the mapping, RRUFF benchmarks, subgroup graph, and four ViT checkpoints first released in our earlier work~\citep{baggett2026attention}. Using these checkpoints, an in-house 1D residual CNN, and six architectures trained under the XRDBench recipe~\citep{simxrd2025}, we show that evaluation design changes what conclusions are justified: standard paired tests point in different directions depending on whether one evaluates Top-1 or Top-5; paired Top-$k$ tests do not resolve differences among the three leading XRDBench models, while hierarchy-aware analysis describes error distance and direction across all six; an input-blind frequency predictor matches the models on aggregate Top-$k$, yet a label-permutation test still detects pattern-specific evidence; and stratified analysis exposes a label-entropy confound that aggregate accuracy hides. All mapping tables, benchmark code, scripts, and checkpoints are released as a shared infrastructure for further PXRD-ML evaluation.
 
\end{abstract}

\section{Introduction}
 
Although powder X-ray diffraction (PXRD) symmetry-classification models are being trained at scale, the field of ML for PXRD has no agreed-upon evaluation standard. Different studies use incompatible subsets of available experimental archives (namely RRUFF~\citep{rruff2019}), different data filters, target taxonomies, and metrics, and rarely release runnable inference artifacts~\citep{park2017classification,Lolla2022SemiSupervisedCrystal,schopmans_neural_2023,10887635,simxrd2025}. Reported accuracy conflates model progress with benchmark and target-design choices~\citep{schopmans_neural_2023,simxrd2025}. We identify three systematic failures in current PXRD-ML evaluation:

\begin{itemize}[leftmargin=*,nosep]
\item \textbf{Wrong target.} Powder diffraction cannot distinguish space groups, the common classification target, within an extinction group, so 230-way classification penalizes a model for conflating patterns it cannot tell apart.
\item \textbf{Non-reproducible benchmarks.} Hand-filtered RRUFF subsets remove the hardest cases, exclusion criteria vary between papers without algorithmic specification, and duplicates distort accuracy.
\item \textbf{Flat metrics and missing baselines.} Top-$k$ accuracy treats a model predicting a nearby subgroup the same as predicting an unrelated symmetry, and without a prior-only baseline one cannot tell whether accuracy comes from diffraction evidence or from reproducing the label distribution.
\end{itemize}

These failures compound: target taxonomy, simulator realism, and post-hoc calibration all affect whether real-data behavior reflects diffraction evidence or dataset bias, and without released checkpoints third parties cannot rescore prior work on a common test set. Neither recent studies \citet{schopmans_neural_2023} nor SimXRD~\citep{simxrd2025} releases such artifacts.

\paragraph{Scope and contributions.} This paper contributes \emph{evaluation infrastructure and standards}, not a new PXRD architecture, to make future PXRD-ML claims more comparable, physically meaningful, and auditable.
\begin{enumerate}[leftmargin=*,nosep]
\item A curated \textbf{extinction-group mapping table} (230 $\rightarrow$ 99), first released with \citet{baggett2026attention}, provided here as a processed, verified machine-readable artifact. (\S\ref{sec:mapping})
\item A \textbf{benchmark-construction method} specified as code, reproducible from a frozen snapshot, that yields reproducible benchmarks \textbf{RRUFF-473} and \textbf{RRUFF-325} from \citet{baggett2026attention}, and additionally curated over the \textbf{opXRD} dataset as a case example. (\S\ref{sec:benchmarks})
\item \textbf{Evaluation tools}: prior-only frequency baselines with deployable and oracle variants, a label-permutation test, the hierarchy-aware DAG metric from our earlier work~\citep{baggett2026attention} supplemented with exact nulls, and stratified reporting. (\S\ref{sec:metrics})
\item A \textbf{demonstration} on four ViT checkpoints~\citep{baggett2026attention}, a 1D CNN, and six XRDBench architectures~\citep{simxrd2025} that these choices change rankings and reveal confounds. (\S\ref{sec:revealed})
\end{enumerate}

\paragraph{Relationship to Baggett et al.} This paper builds off our prior work \citet{baggett2026attention}, which first introduced and publicly released the mapping, RRUFF-473 and RRUFF-325, the condensed subgroup graph, and the four ViT checkpoints; each is credited where it is introduced. The two papers ask different questions: Baggett et al.\ ask whether a physics-informed model can read symmetry from a powder pattern, whereas this paper asks what it takes to know whether any model can. To our knowledge those checkpoints remain the only PXRD-ML transformer models with a public inference pipeline that can be rescored under this protocol.

\paragraph{Released artifacts.}  The mapping table, both benchmarks with curation code and record lists, the evaluation scripts, and the case-study checkpoints are public; components reused from \citet{baggett2026attention} carry their provenance. Appendix~\ref{app:artifacts} lists the repositories and the full inventory.
 
\begin{center}
\href{https://github.com/scattering/InformedXRD}{\faGithub~Code}\hspace{2.5em}
\href{https://huggingface.co/datasets/informedxrd/informedxrd-benchmarks}{\hficon~Benchmarks}\hspace{2.5em}
\href{https://huggingface.co/informedxrd/informedxrd-checkpoints}{\hficon~Checkpoints}
\end{center}

\section{Background}
\label{sec:background}

\paragraph{Crystal structure and powder diffraction.} A crystal structure is a periodic arrangement of atoms, classified by its internal symmetry into 230 \textit{space groups}. Powder X-ray diffraction (PXRD) experiments produce a 1D intensity trace whose peak positions and relative intensities encode symmetry and atomic arrangement. The relevant fact for classification is a built-in resolution limit: symmetry elements such as glide planes and screw axes produce \textit{systematic absences}---reflections forbidden by symmetry and therefore missing from the pattern---but \textbf{different space groups can share identical absence sets}, making them \textbf{experimentally indistinguishable from a powder pattern alone}. Friedel's law further removes the distinction between centrosymmetric and noncentrosymmetric structures (with and without inversion symmetry) when anomalous scattering is negligible. A benchmark scoring models beyond these limits asks for information the measurement does not contain.

\paragraph{Extinction groups.} \label{sec:extgroupsdisscussion}The equivalence classes of space groups that share identical systematic absences are called \textit{extinction groups}, which reduce 230 space groups to 99 extinction groups~\citep{int_tables_a} (Appendix~\ref{app:mapping_build} gives the construction). They are the finest evaluative label space PXRD can justify, so scoring in 230-way space-group space risks penalizing a physically correct prediction because the measurement cannot distinguish two classes; \S\ref{sec:mapping} describes the penalty of using space groups. One hexagonal/trigonal pair of extinction classes is merged only in the condensed subgroup graph used for the hierarchy metric (\S\ref{sec:dagmetric}) scoring, though not in standard label scoring (see Appendix~\ref{app:hex_trig} for more).

\paragraph{Subgroup hierarchy.} Extinction groups are not independent classes but are related through a crystallographic subgroup hierarchy: each connects to others through \textit{translationengleiche} ($t$-) subgroup relations~\citep{int_tables_a1}, which preserve the translation lattice while removing point-group operations such as rotations, reflections, and inversion. These form a directed acyclic graph with edges from higher to lower symmetry; the condensed $t$-subgroup graph provides a reproducible first-order notion of error severity with clean ancestor/descendant semantics.

\paragraph{ML for PXRD classification.} Labeled experimental patterns are scarce, so PXRD models are trained almost entirely on simulated data. Nearly every study computes its training patterns from structures in a crystal-structure database~\citep{park2017classification,vecsei2019neural,oviedo2019fast,suzuki2020symmetry,zaloga2020crystal,Lolla2022SemiSupervisedCrystal,lee2022pxrdallyouneed,salgado2023automated,corriero2023crystalmela,ozaki2025spgvolnet,andrejevic2026alphadiffract,simxrd2025,wang2025extinction}, most often the Inorganic Crystal Structure Database (ICSD)~\citep{zagorac2019icsd}, the Crystallography Open Database (COD)~\citep{grazulis2009cod}, or the Materials Project~\citep{jain2013materialsproject}, and usually augments them with noise, backgrounds, peak broadening, or preferred orientation; a few studies also perturb those structures or generate synthetic crystals of their own~\citep{simxrd2025,szymanski2021probabilistic,wang2025extinction,schopmans_neural_2023,baggett2026attention}, and only rarely do measured patterns enter training~\citep{oviedo2019fast,salgado2023automated}. Measured patterns are instead reserved for evaluation, where the common source is RRUFF~\citep{rruff2019}, a public collection of experimental powder diffraction patterns with a primary mineral phase named for every entry; but each study evaluates on its own hand-selected subset of it, under its own target definition and preprocessing. Open experimental archives such as opXRD~\citep{hollarek2025opxrd} are beginning to expand the available real data (\S\ref{sec:opxrd_instance}). \citet{simxrd2025} reported strong transfer to RRUFF from their SimXRD-4M simulator and XRDBench recipe. Our earlier work~\citep{baggett2026attention} contributed one of the few released PXRD-ML inference pipelines that can actually be rescored under a common target and benchmark, which is why we use those checkpoints as case studies in \S\ref{sec:revealed}.

\section{Evaluating on Extinction Groups and the Mapping Table}
\label{sec:mapping}
Scoring in space-group space understates what a model has learned. Besides \citet{baggett2026attention}, almost all previous studies use space groups as the label space which PXRD cammot resolve (see  \S\ref{sec:background}). To measure the cost, we ran a matched, convergence-controlled comparison with the training pipeline and 2.0M-pattern synthetic dataset of \citet{baggett2026attention}. The same 230-way space-group model, on the same predictions, reaches 15.17\% Top-1 scored in its own space and 18.46\% once its predictions are mapped into the 99-way extinction-group space. Training directly on extinction groups adds less: the direct model reaches 19.32\% at full scale, an advantage of +0.86 \%. Yet this gap widens to +2.77 \% at one-eighth training scale, so direct supervision is primarily a sample-efficiency advantage that the space-group model recovers implicitly as the corpus grows. The full comparison, including Top-5, the intermediate scales, and the RRUFF transfer check, is in Appendix~\ref{app:target_effect}.

Acting on that choice requires a machine-readable, canonicalized label space. The underlying tabulation is public---the extinction-symbol table distributed with EXPO~\citep{altomare_expo}, derived from \textit{International Tables for Crystallography}~\citep{int_tables_a} and first released in machine-usable form with \citet{baggett2026attention}---and we parse, canonicalize, and invert it into a 230-row lookup cross-checked against \textit{International Tables} (Appendix~\ref{app:mapping_build}). The contribution is the processed, verified artifact, released as JSON and CSV, under which any 230-way model can be rescored post hoc and misalignment is auditable.

\section{Standardized Benchmark Construction} \label{sec:benchmarks}
The construction method is defined over the label space and the measurement grid, not over any one archive. This section states the method (\S\ref{sec:construction}) and applies it to two archives: RRUFF (\S\ref{sec:rruff_instance}) yielding RRUFF-473 and RRUFF-325, on which \S\ref{sec:revealed} evaluates models, and opXRD (\S\ref{sec:opxrd_instance}), a small example yielding opXRD-98, evaluated in Appendix~\ref{app:opxrd}.

\subsection{Construction Method}
\label{sec:construction}
A source archive isn't always a usable benchmark at first: it may contain multiple scans per material, incomplete or multiphase labels, mixed wavelengths, and exact duplicates. The construction method turns it into a fixed evaluation set whose inclusion rules are specified as code and reproducable from a frozen snapshot:
\begin{enumerate}[leftmargin=*,nosep]
\item \textbf{Single assignable label.} Retain records with one phase whose space group is stored or derivable, mapped to the label space of \S\ref{sec:mapping}; multiphase records without a primary phase are excluded.
\item \textbf{Wavelength and grid gate.} Select scans at the target wavelength, require complete angular support at native resolution over the evaluation window, and resample onto a standard $2\theta$ grid.
\item \textbf{Structural grouping.} Assign each record an archive-specific structural-family label, used both to group near-identical records and to stratify results.
\item \textbf{Family coherence.} Keep only families internally consistent under within-family diagnostics, separating distinct variants and admitting singletons only when they match the family core.
\item \textbf{Exact-pattern deduplication}, conflict-aware: byte-identical patterns reduce to one representative, and groups carrying conflicting labels are excluded.
\item \textbf{Per-pattern strata.} Attach nuisance-fit severity from classical Rietveld fitting under fixed $R_\mathrm{wp}$ cutoffs (\S\ref{sec:stratified}), and record any measurement stratum that varies within the archive.
\item \textbf{Freeze and audit.} Release snapshot-dated identifiers, curation code, and a threshold sweep.
\end{enumerate}
Steps 1, 2, and 7 are identical for every archive; steps 3--6 can require a different instantiation. Figure~\ref{fig:pipelines} records both instantiations side by side; the full step descriptions are in Appendix~\ref{app:method}, and the thresholds and gate counts in Appendix~\ref{app:rruff_details} (RRUFF) and Appendix~\ref{app:opxrd} (opXRD). Sets from different archives are reported separately, never pooled.

\begin{figure}[t]
\centering
\includegraphics[width=\linewidth]{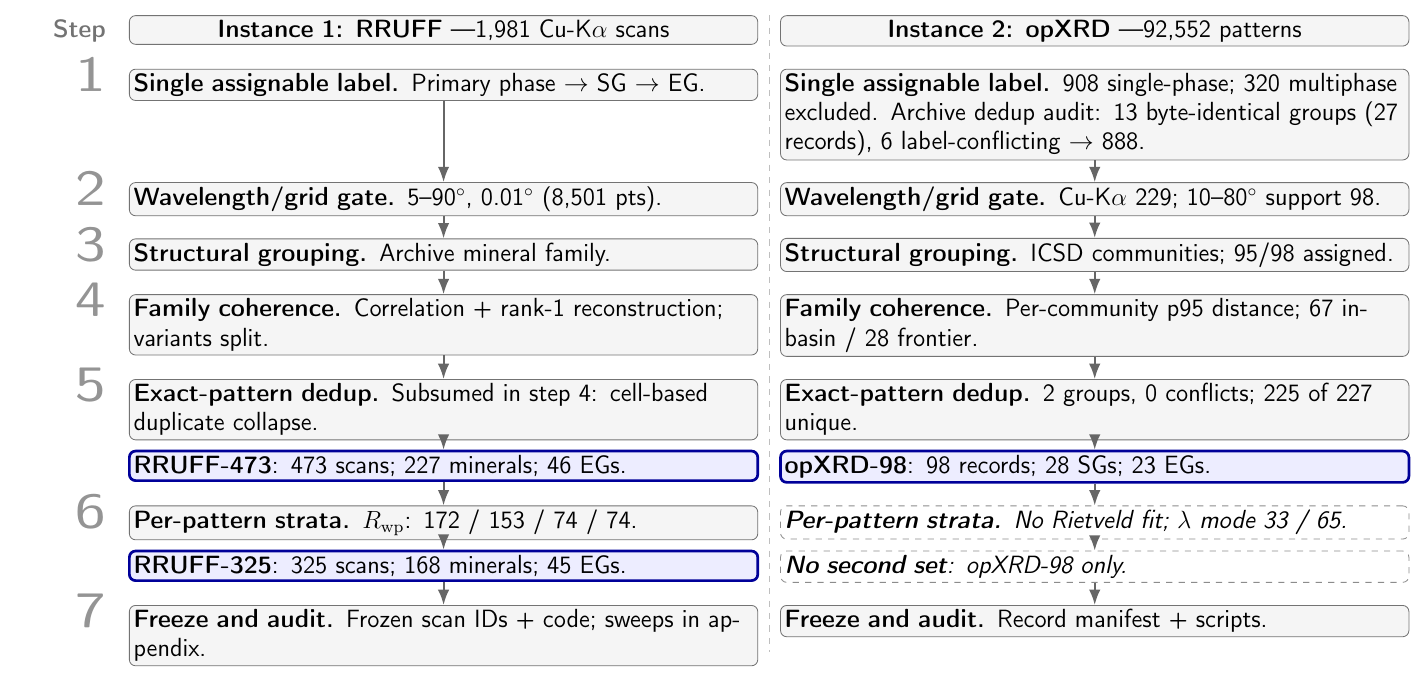}
\caption{The construction method of \S\ref{sec:construction} instantiated on two archives. Left: RRUFF, yielding the frozen RRUFF-473 and RRUFF-325 benchmarks with their nuisance-fit strata. Right: opXRD, a small example yielding opXRD-98. Rows are the steps of the method aligned across the two archives, with dashed boxes marking omitted steps and blue boxes marking the released evaluation sets. Per-step detail---thresholds, diagnostics, and the full gate cascade---is in Appendix~\ref{app:rruff_details} (RRUFF) and Appendix~\ref{app:opxrd} (opXRD).}
\label{fig:pipelines}
\end{figure}

\subsection{Instance 1: RRUFF-473 and RRUFF-325}
\label{sec:rruff_instance}
RRUFF~\citep{rruff2019} is one of few public sources of experimental PXRD with primary-phase labels, so RRUFF-473 and RRUFF-325, introduced in our earlier work~\citep{baggett2026attention}, are the primary benchmarks here: real, singly labeled, reproducible from a frozen snapshot. Family-coherence checks reduce the 1{,}981-scan Cu-K$\alpha$ pool to RRUFF-473; families that fail these checks are excluded because they would make the benchmark ambiguous before any model is evaluated. Nuisance-fit cutoffs applied to the frozen RRUFF-473 then give RRUFF-325. Third parties should be able to reconstruct what was evaluated, why those scans were retained, and how composition affects the resulting claims; the full pipeline, released defaults, and threshold-sensitivity results are in Appendix~\ref{app:rruff_details}.

\paragraph{Reproducibility, coverage, and intended use.} Released scan-ID lists, snapshot date, curation code, and thresholds reproduce both sets exactly, and RRUFF-325 is a deterministic subset of RRUFF-473. Both are fixed evaluation sets, not training corpora, covering 46 and 45 of the 99 extinction groups (Table~\ref{tab:prior_studies_comp}), so claims do not extend to unrepresented groups (Appendix~\ref{app:rruff_details} gives the coverage rationale). Report both---RRUFF-473 for real-data robustness, RRUFF-325 for cleaner error structure---to avoid overfitting to one slice.

\subsection{Instance 2: a Small Example, opXRD-98}
\label{sec:opxrd_instance}
opXRD~\citep{hollarek2025opxrd}, an open archive of 92{,}552 experimental patterns from multiple laboratories, is the second instance and tests whether the method transfers beyond RRUFF. Its scale benefits representation learning, but the one-pattern, one-label protocol yields only a small example, opXRD-98, of 98 records; the taxonomy and both readouts transfer without modification, but the result is a stress test rather than a benchmark, and we hope it encourages larger labeled archives. The exact record list and processing scripts are released; the gate table, the reasons the benchmark was not built on opXRD or merged with RRUFF, and the evaluation results are in Appendix~\ref{app:opxrd}. One step is instantiated differently: RRUFF's mineral family does two jobs---grouping structurally near-identical records and stratifying results---and opXRD carries no such label, so both are done from structure alone by projecting its records into the frozen ICSD structural map of \citet{nguyen2026icsdmap} (Appendix~\ref{app:opxrd}).

\section{Evaluation Protocol}
\label{sec:metrics}
We propose three evaluation tools that should accompany Top-$k$ accuracy in any PXRD-ML evaluation. The first (\S\ref{sec:priorbaseline}) pairs input-blind prior-only baselines with a label-permutation test: the baselines establish how much Top-$k$ accuracy follows from label frequency alone, and the permutation test asks whether a model's rankings depend on the individual diffraction pattern. The second (\S\ref{sec:dagmetric}) replaces a single scalar with a hierarchy-aware characterization of \emph{how} the model is wrong. The third (\S\ref{sec:stratified}) partitions performance along axes that aggregate accuracy collapses.

\subsection{Prior-Only Frequency Baseline and Label Permutation Test}
\label{sec:priorbaseline}
Crystallographic training distributions are heavily imbalanced (Figure~\ref{fig:icsd_distribution}, Appendix~\ref{app:artifacts}), so a model can score well on real data by reproducing the label distribution rather than by reading the diffraction pattern. The prior-only baseline detects this by ranking the extinction groups by their training frequency and predicting that same fixed Top-$k$ list for every pattern, ignoring the input. We use the aggregate prior of the case-study calibration pipeline, estimated from its synthetic training corpus and released with the paper (Appendix~\ref{app:artifacts}). We also report each benchmark's own empirical label frequencies as an oracle prior-only control: computed from the test labels and serving as an upper bound on what any frequency-only predictor can achieve on that fixed set without inspecting the pattern. \S\ref{sec:evidence} applies both controls to the case-study models, and \S\ref{sec:paradox} uses the oracle control per severity stratum.

\paragraph{Label-permutation test.} The label-permutation test asks what the prior-only baselines cannot: whether a model's rankings depend on the individual diffraction pattern. Hold the model's Top-$k$ candidate set fixed for every diffraction pattern and randomly reassign the benchmark labels among patterns. This preserves both the benchmark label frequencies and the model's tendency to predict common groups, while destroying only the association between a particular diffraction pattern and its correct label. The shuffled-label mean is what the same candidate sets would score on random scans from the same benchmark, and a one-sided permutation $p$-value compares the observed Top-$k$ (or mean true-label rank) against it; we use 10{,}000 permutations. Together the two controls and the test give a three-part reading of any model: whether it clears the deployable prior, where it sits relative to the oracle prior-only control, and whether its pattern-specific rankings retain a detectable association with the correct scans.

\subsection{Hierarchy-Aware DAG Metric}
\label{sec:dagmetric}
Top-1 accuracy treats every misclassification as an equal failure, but from a crystallographic view, errors have structure. Misclassifying into a maximal subgroup---one edge away on the symmetry hierarchy---is a nearby mistake within the same family of symmetries; predicting an extinction group from a completely different crystal-system branch is a much more severe error.

We formalize this as graph distance on the condensed \textit{translationengleiche} ($t$-subgroup) DAG released with \citet{baggett2026attention}, which we adopt unchanged. \emph{Nodes}: the 99 extinction groups, with the hexagonal/trigonal extinction-class pair $P\text{-}\text{-}c$ / $P\text{-}c\text{-}$ condensed into a single node (see the rationale in Appendix~\ref{app:hex_trig}), giving 98 nodes. \emph{Edges}: $t$-subgroup relations, directed from higher to lower symmetry. We use unweighted hop distance on the $t$-subgroup graph alone (\S\ref{sec:background}); weighted or mixed extensions are plausible but would add standardization choices, so the present metric is an auditable default rather than the only possible severity geometry.

For each wrong Top-1 prediction we compute three quantities: distance, direction, and locality. \textbf{Error distance} is the shortest-path distance between the predicted and true extinction group on the DAG, reported as the mean and the full distribution. \textbf{Directionality} records whether the prediction is a \textit{descendant} of the truth (a subgroup: the model fell to lower symmetry), an \textit{ancestor} (a supergroup: it asserted higher symmetry), or a \textit{branch jump} (a different crystal-system family). \textbf{Locality fraction} is the share of wrong predictions within two hops. 

Each quantity has a physical reading, and together they separate conservative errors from severe ones. Short distances with a descendant bias are the conservative error mode: noise erases evidence of high symmetry first, so falling to a nearby lower-symmetry class is the expected failure. Long distances with an ancestor bias are the opposite: the model asserts symmetry the pattern contradicts, often by leaning on the label prior. A high branch-jump rate is the most severe mode: the model leaves the local symmetry neighborhood, typically when strong preferred orientation suppresses a diagnostic peak family and the surviving pattern resembles a different branch. Distance and direction must be read together: a short mean distance can arise from genuinely local confusions or from a bias toward low-symmetry nodes, which are near-descendants of many groups.

\paragraph{Reference distributions.} A hierarchy-aware summary is interpretable only against what an uninformed predictor would score on the same graph, so we report it against four references (values in Table~\ref{tab:dag_nulls}, Appendix~\ref{app:dag_nulls}). The \emph{uniform null} enumerates, for every benchmark true label, all 98 incorrect extinction-group predictions with equal weight and measures the shortest distance on the same condensed graph; it is an exact expectation rather than a sample. Because models differ in which cases they get wrong, the \emph{model-matched null} repeats the computation over each model's own error set. The \emph{frequency-matched null} draws the wrong label from the benchmark's own extinction-group marginal and is the harder reference. The fourth reference, always naming the most frequent group, shows why distance must be read with direction (\S\ref{sec:disagree}). All four are reported alongside the observed summaries. \S\ref{sec:disagree} reads the ViT checkpoints and the CNN control against them, \S\ref{sec:xrdbench} reads six independently trained architectures the same way, and Appendix~\ref{app:opxrd} does so for opXRD-98.

\paragraph{Relation to prior hierarchical metrics.} Distance-based evaluation on a class hierarchy is a long-standing family of hierarchical-classification metrics~\citep{silla2011survey,cao2024consistent}; we do not claim hop-distance-on-a-DAG as a new construction, only its crystallographic instantiation (Appendix~\ref{app:hier_metrics}).

\subsection{Stratified Evaluation}
\label{sec:stratified}

An aggregate accuracy number over a benchmark collapses two distinct axes of variation: the intrinsic difficulty of the classification problem within a subset of the data, and the number of distinct labels to choose between in that subset. Stratified evaluation partitions the benchmark along both axes and reports per-stratum accuracy alongside the aggregate.

\textbf{Data quality axis.} Each RRUFF-473 pattern carries a nuisance-fit severity label---recoverable, usable, poor, or catastrophic---based on how well classical Rietveld profile fitting, the standard full-pattern least-squares refinement workflow in powder crystallography, handles it. Here ``severity'' means the severity of the residual mismatch after a standard classical fit, not the importance of the sample itself: usable patterns fit relatively cleanly, recoverable patterns remain interpretable but noticeably degraded, poor patterns retain substantial mismatch, and catastrophic patterns fail badly enough to act as stress tests.

\textbf{Label-space entropy axis.} Each stratum also contains a different number of unique extinction groups, and a narrower label distribution is intrinsically easier to classify, independent of data quality (\S\ref{sec:paradox} gives the RRUFF-473 counts). Reporting aggregate accuracy without the entropy decomposition conflates ``the model does well on this stratum'' with ``this stratum has fewer classes to confuse.''

The protocol therefore reports aggregate Top-$k$ alongside per-stratum Top-$k$, the number of unique extinction groups per stratum, and the entropy-reweighted per-stratum Top-$k$ that adjusts for label-space concentration. When a per-stratum effect persists after reweighting, it is a property of the model's interaction with the data; when it disappears, it was a property of the label distribution. \S\ref{sec:paradox} applies this decomposition to the severity strata of RRUFF-473, where it changes the interpretation of an apparent quality gap.

\section{What Standard Evaluation Hides}
\label{sec:revealed}
We apply the framework to three model families: four publicly released ViT checkpoints from \citet{baggett2026attention} (no-preferred-orientation, preferred-orientation-only, final mixed-data calibrated, and legacy uncalibrated); an in-house 1D residual CNN control (Appendix~\ref{app:cnn}); and six independently developed architectures---CNN10, CNN11, PatchTST, BiLSTM, BiGRU, and a generic Transformer---trained under the XRDBench recipe of SimXRD~\citep{simxrd2025} with 230-way outputs, collapsed into the 99-way space with the \S\ref{sec:mapping} mapping and scored on identical RRUFF-473 and RRUFF-325 inputs (\S\ref{sec:rruff_instance}). Five findings follow, each about how the field reports results: Top-1 and Top-5 rank the same checkpoints in opposite directions, and the hierarchy-aware metric needs its own null (\S\ref{sec:disagree}); paired Top-$k$ tests do not resolve differences among CNN10, CNN11, and PatchTST, while hierarchy-aware analysis describes error distance and direction across all six architectures (\S\ref{sec:xrdbench}); an input-blind predictor matches the models on aggregate Top-$k$, and only a permutation test recovers the pattern-specific evidence (\S\ref{sec:evidence}); and classical fit quality does not predict neural difficulty once label-space concentration is controlled (\S\ref{sec:paradox}). The small opXRD-98 example (\S\ref{sec:opxrd_instance}) is scored under the same protocol in Appendix~\ref{app:opxrd}.

\subsection{Flat Accuracy and Hierarchy-Aware Severity Disagree on Model Ranking}
\label{sec:disagree}

On RRUFF-325, flat Top-1 and hierarchy-aware severity rank the five models differently (Table~\ref{tab:rank_disagreement}; Figure~\ref{fig:metric_disagreement} in Appendix~\ref{app:dag_nulls} shows same reversal): the model one would pick under flat Top-1, the preferred-orientation-only ViT checkpoint, is not the model one would pick under mean DAG distance, the final mixed-data calibrated checkpoint.

\begin{table}[!htbp]
\centering
\caption{Five-model comparison on RRUFF-325: Top-1, $\le 2$-hop locality, mean DAG distance, and branch-jump fraction, with the implied ranks under flat Top-1 versus mean DAG distance. The two ranking columns disagree on the preferred-orientation-only and final mixed-data calibrated ViT checkpoints; the 1D CNN baseline is set apart from the ViT family by its 53.8\% branch-jump rate. Hierarchy columns are over wrong Top-1 predictions; matched nulls for the preferred-orientation-only and final mixed-data calibrated ViTs and the CNN are given in Table~\ref{tab:dag_nulls}. Boldface marks the best value in each column and does not denote statistical significance.}
\label{tab:rank_disagreement}
\scriptsize
\setlength{\tabcolsep}{4pt}
\renewcommand{\arraystretch}{1.1}
\begin{tabular}{lcccccc}
\toprule
Model & Top-1 (\%) & $\le$2 hops (\%) & Mean dist. ($\downarrow$) & Branch (\%, $\downarrow$) & Top-1 rank & Dist. rank \\
\midrule
ViT: no preferred orientation      & 9.54  & 38.4  & 2.72 & \textbf{25.8} & 3 & 3 \\
ViT: preferred-orientation only    & \textbf{13.54} & \textbf{53.0}  & 2.51 & 34.1 & \textbf{1} & 2 \\
ViT: final mixed-data calibrated   & 10.46 & 51.55 & \textbf{2.46} & 32.6 & 2 & \textbf{1} \\
1D residual CNN        & 6.77  & 28.05 & 2.92 & 53.8 & 4 & 4 \\
ViT: legacy uncalibrated & 4.92  & 17.8  & 3.50 & 56.4 & 5 & 5 \\
\bottomrule
\end{tabular}
\end{table}

Standard paired tests make the disagreement concrete: on the same 325 examples, Top-1 and Top-5 separate the ViT pair in opposite directions, while the hierarchy-aware summaries do not separate it at all. Exact McNemar tests give the preferred-orientation-only checkpoint the win at Top-1 (13.54\% vs.\ 10.46\%, $p=0.002$) and the final calibrated checkpoint the win at Top-5 (43.69\% vs.\ 40.92\%, $p=0.012$). Paired permutation tests on hierarchy-aware summaries do not separate the pair (mean severity-burden $\Delta=-0.034$, $p=0.58$); bootstrap CIs for the final calibrated checkpoint are reported in Appendix~\ref{app:ci}. Even under standard paired testing, then, the same predictions support different conclusions depending on whether the benchmark asks for exact rank-1 identification or broader candidate recovery---and the hierarchy-aware metrics still expose distinct notions of error structure (the preferred-orientation-only checkpoint retains the highest $\leq 2$-hop locality at 53.0\%, while the final calibrated checkpoint achieves the tightest mean distance and best Top-5 behavior).

The framework also distinguishes architecture families, not just training variants. The 1D residual CNN control (Appendix~\ref{app:cnn}), evaluated at its epoch-6 checkpoint on RRUFF-325, reaches 6.77\% Top-1---below the deployable prior-only baseline---but the more informative observation is structural: only 28.05\% of its wrong predictions fall within $\leq 2$ hops, and 53.8\% of its mistakes are branch jumps, versus 32.6\% for the final mixed-data calibrated ViT checkpoint (Table~\ref{tab:rank_disagreement})---a qualitatively different failure regime that flat Top-1 alone would suppress. Because an undertrained control could inflate the transformer family's advantage, we tested convergence directly: resuming the same lineage for 40 further epochs and selecting on held-out synthetic validation loss alone, epoch 46 reaches the lowest validation loss yet is worse on both real benchmarks at every $k$ and less local on the DAG (Table~\ref{tab:cnn_convergence}). The submitted checkpoint was therefore the more favorable one for the CNN, and synthetic validation improving while both real benchmarks degrade indicates simulation-to-experiment domain shift rather than an intrinsic CNN limitation. The six XRDBench architectures of \S\ref{sec:xrdbench} carry the comparative claims; the CNN stands as one fixed reference among several.

\paragraph{Reading the hierarchy metric against its nulls.} Against the reference distributions of \S\ref{sec:dagmetric} (Table~\ref{tab:dag_nulls}), the final mixed-data ViT is genuinely more local than chance while the CNN control is not. On RRUFF-325 the ViT is well inside its model-matched null on both distance and locality, is strongly biased toward lower-symmetry descendants, and also clears the harder frequency-matched null on both measures. The CNN control sits at its matched null on distance, with a much smaller directional shift. One caution follows. Always naming the most frequent group scores 2.116 and 63.4\% on RRUFF-325, better than any model on distance alone, because low-symmetry nodes are near-descendants of many groups; a low mean distance with a strong descendant bias is therefore a different finding from a low mean distance with a branch-heavy error set. The preferred-orientation-only checkpoint, which wins flat Top-1, also sits well inside its own matched null (Table~\ref{tab:dag_nulls}), so the ViT pair is separated by flat metrics but not by hierarchy-aware ones.

\subsection{Accuracy and Error Structure Across Independent Architectures}
\label{sec:xrdbench}

Paired tests do not resolve Top-1, Top-3, or Top-5 differences among CNN10, CNN11, and PatchTST on either RRUFF benchmark. Across the six architectures, hierarchy-aware analysis reveals differences in error distance and direction that aggregate accuracy does not describe (Table~\ref{tab:xrdbench}). We trained the six models under the XRDBench recipe because four checkpoints from one ViT family are not enough to establish that evaluation design changes conclusions across architectures, and scored them on identical RRUFF-473 and RRUFF-325 inputs. Exact two-sided McNemar tests on every pair among CNN10, PatchTST, and CNN11 fail at Top-1, Top-3, and Top-5 on both benchmarks, the smallest raw $p$ being $0.124$ (Holm-adjusted $0.371$, for PatchTST versus CNN11 at RRUFF-325 Top-3), and every 10{,}000-replicate paired-bootstrap interval includes zero; the largest tested difference is $+4.0$ points with 95\% CI $[-0.6, 8.6]$.

The hierarchy-aware summaries provide a different view of the errors. PatchTST, CNN11, and CNN10 are clearly more local than their own model-matched nulls; BiLSTM and BiGRU sit close to theirs; and the generic Transformer is markedly worse than its own null (16.71\% against 32.54\% within two hops on RRUFF-473; 16.88\% on RRUFF-325) while directing 95.8\% of its errors toward lower-symmetry descendants against 9.6\% under its matched null. Because the distance margins narrow against the frequency-matched null---PatchTST's RRUFF-473 result (2.688, 40.10\%) does not clear it (2.692, 42.15\%)---we treat the directional contrast as the firmer result. Together with \S\ref{sec:disagree}, these results show how hierarchy-aware analysis complements accuracy by describing the distance and direction of errors. Neither readout dominates, and what a benchmark can resolve depends on the evaluation design, not only on the models.

\begin{table}[!htbp]
\centering
\caption{Six XRDBench architectures rescored in extinction-group space. Models were trained on the unchanged public SimXRD \texttt{train\_1} contract~\citep{simxrd2025} and evaluated on RRUFF patterns exactly decimated onto that grid (contract specification in Appendix~\ref{app:artifacts}). Hierarchy columns are over wrong Top-1 predictions and give observed / model-matched null (\S\ref{sec:dagmetric}); on RRUFF-325 matched nulls were computed for the two strongest models. No pair among CNN10, PatchTST, and CNN11 is separable by exact McNemar test at Top-1, Top-3, or Top-5 on either benchmark (smallest raw $p=0.124$). Boldface marks the best value per column, not statistical significance.}
\label{tab:xrdbench}
\scriptsize
\setlength{\tabcolsep}{3.5pt}
\begin{tabular}{lcccccc}
\toprule
 & \multicolumn{3}{c}{RRUFF-473} & \multicolumn{3}{c}{RRUFF-325} \\
\cmidrule(lr){2-4}\cmidrule(lr){5-7}
Model & Top-1 / 5 (\%) & $\le$2 hops obs./null & Mean dist. obs./null & Top-1 / 5 (\%) & $\le$2 hops obs./null & Mean dist. obs./null \\
\midrule
CNN10       & \textbf{12.47} / 45.67 & 37.68 / 31.43 & 2.797 / 2.910 & \textbf{9.85} / 39.08 & 35.15 / 31.42 & 2.877 / 2.901 \\
PatchTST    & \textbf{12.47} / 45.45 & \textbf{40.10} / 30.01 & \textbf{2.688} / 2.969 & 9.23 / \textbf{40.62} & \textbf{37.29} / 30.84 & \textbf{2.783} / 2.927 \\
BiLSTM      & 8.67 / 30.87  & 34.49 / 31.26 & 2.894 / 2.929 & 4.92 / 24.00 & 34.95 / --- & 2.903 / --- \\
CNN11       & 11.63 / \textbf{45.88} & 38.52 / 31.09 & 2.749 / 2.919 & 8.31 / 40.31 & 36.24 / --- & 2.842 / --- \\
BiGRU       & 10.57 / 33.62 & 32.86 / 31.75 & 2.877 / 2.901 & 6.77 / 25.85 & 31.68 / --- & 2.921 / --- \\
Transformer & 10.15 / 40.38 & 16.71 / 32.54 & 3.386 / 2.864 & 5.23 / 33.54 & 16.88 / --- & 3.461 / --- \\
\bottomrule
\end{tabular}
\end{table}

\subsection{Prior-Only Baselines and the Permutation Test: Where the Evidence Is}
\label{sec:evidence}

The final calibrated model clears the deployable prior of \S\ref{sec:priorbaseline} but does not exceed the oracle prior-only control, and the two controls must be kept apart. The \emph{deployable} prior reaches 9.7\% Top-1 and 27.2\% Top-5 on extinction-group labels derived from real RRUFF patterns. The \emph{oracle} prior gives the highest accuracy attainable by a fixed input-blind ranking on each benchmark, at 12.9/51.4\% on RRUFF-473 and 12.6/43.7\% on RRUFF-325 (Table~\ref{tab:benchmark_prior_controls}). The final mixed-data calibrated ViT checkpoint reaches 9.94/50.74\% on RRUFF-473 and 10.46/43.69\% on RRUFF-325: above the deployable prior at Top-5 by a wide margin, below the oracle control except for a tie at RRUFF-325 Top-5. Stated plainly, an input-blind predictor that knew the benchmark's label histogram would match or exceed this model's aggregate Top-$k$, and the comparison with the deployable prior does not by itself establish that the model uses the diffraction pattern. A model using the diffraction pattern can outperform the oracle prior-only control. On RRUFF-325 the preferred-orientation-only ViT does so at Top-1, reaching 13.54\% against the oracle's 12.6\%. On opXRD-98 (Appendix~\ref{app:opxrd}) the same control is even more decisive: an input-blind predictor using opXRD-98's own label marginal scores 22.45/66.33\%, above every one of the six XRDBench models.

The label-permutation test of \S\ref{sec:priorbaseline}, applied to each pattern's Top-5 candidate set, answers the narrower question, whether the model's rankings depend on the individual diffraction pattern, in the affirmative but by a small margin. Observed RRUFF-473 Top-5 is 50.74\% against a shuffled-label mean of 48.52\% (+2.22 points, one-sided $p=0.0131$). On RRUFF-325, observed Top-5 is 43.69\% against 41.48\% after shuffling (+2.21 points, $p=0.0282$), and the mean rank of the true label improves from the shuffled 13.71 to 13.43 ($p=0.0300$); the RRUFF-473 mean-rank test is weaker ($p=0.061$). The tie with the oracle at RRUFF-325 Top-5 is instructive: model and fixed prior are each Top-5-correct on 142 of 325 patterns, but the sets differ (132 shared, 10 model-only, 10 prior-only), which is exactly the pattern-specific component the permutation test isolates. The summary is three-part: the model clears the deployable prior at Top-5, matches or falls below the oracle prior-only control in aggregate, and its pattern-specific rankings retain a small but statistically detectable association with the correct scans.

\subsection{The Catastrophic Paradox as a Decomposition Exercise}
\label{sec:paradox}

Applying stratified evaluation to RRUFF-473 reveals a counterintuitive observation: classical Rietveld fit quality does not cleanly predict neural classification difficulty. On the final mixed-data calibrated ViT checkpoint, raw Top-1 across the recoverable, usable-or-better, poor, and catastrophic strata is 11.05\%, 9.80\%, 10.81\%, and 6.76\%, respectively. After reweighting each stratum to a common extinction-group distribution, the corresponding values become 13.18\%, 10.61\%, 10.71\%, and 10.80\%. Thus, differences in label-space composition explain about 45\% of the recoverable-versus-catastrophic gap, but not all of it.

\begin{figure}[!htbp]
\centering
\includegraphics[width=\linewidth]{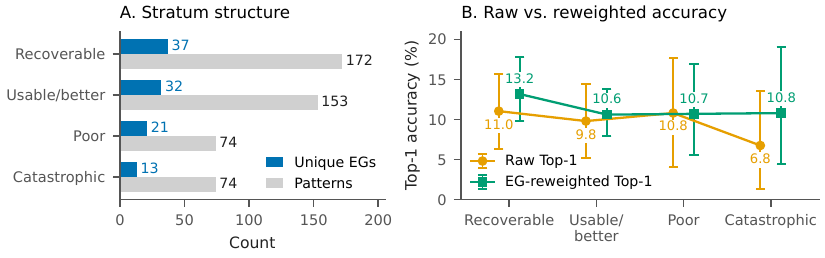}
\caption{Decomposition of the catastrophic paradox on RRUFF-473. Left: the severity strata differ strongly in label-space structure, with the catastrophic stratum spanning far fewer unique extinction groups than the recoverable stratum. Right: classical profile-fit quality and neural classification difficulty do not align monotonically, and extinction-group reweighting removes about 45\% of the recoverable-versus-catastrophic gap.}
\label{fig:catastrophic_paradox}
\end{figure}

The aggregate Top-1 on RRUFF-473 is 9.94\%; under standard reporting that is all a reader sees, yet three mechanisms contribute to it and the aggregate cannot separate them: \emph{label-space concentration} (the catastrophic stratum spans only 13 unique extinction groups to recoverable's 37, so a model can score well with a narrower effective classifier); \emph{intrinsic classification difficulty} (the residual gap after entropy reweighting reflects whatever structure in the data actually matters to the model); and \emph{aggregation bias} (the catastrophic stratum is 74 patterns versus recoverable's 172, so its contribution to the aggregate is weighted accordingly). Only stratified reporting with entropy reweighting separates these: in the final mixed model, reweighting compresses the three harsher strata into a narrow 10.6--10.8\% band, while recoverable remains distinctly higher. The benchmark-matched prior-only ceilings vary sharply across the same strata, from 12.2\%/41.3\% on recoverable to 28.4\%/75.7\% on catastrophic at Top-1/Top-5 (Table~\ref{tab:benchmark_prior_controls}), confirming that label-space concentration differs dramatically across the severity partition. The residual therefore says something specific: recoverable patterns are harder for a classical refinement procedure yet easier for the network than usable-or-better patterns once label-space concentration is controlled---a decoupling that is invisible from aggregate Top-1 (Figure~\ref{fig:catastrophic_paradox}).

\section{Conclusion}
\label{sec:conclusion}

We have presented an evaluation framework for PXRD machine learning addressing three failures: wrong targets, non-reproducible benchmarks, and uninformative metrics. We show that evaluation choices change the conclusions reached (\S\ref{sec:revealed}) and additionally apply our benchmark-construction method to opXRD (Appendix~\ref{app:opxrd}).

\paragraph{Limitations.} RRUFF-473 and RRUFF-325 cover only 46 and 45 of the 99 extinction groups, and we do not infer beyond them; both are restricted to Cu-K$\alpha$ laboratory data and RRUFF's primary-phase labels, and the paired Top-$k$ tests do not resolve differences among CNN10, CNN11, and PatchTST (\S\ref{sec:xrdbench}). The final calibrated model matches or falls below the oracle prior-only control at Top-1 and Top-5, its pattern-specific rankings retaining only a small but detectable association with the correct scans (\S\ref{sec:evidence}). The DAG is unweighted and $t$-only, and because distance margins narrow against the frequency-matched null we treat cross-architecture distance comparisons as suggestive. Stability is checked by threshold sweeps, not alternative refinement workflows, and the RRUFF transfer check of the target-choice effect uses small, domain-shift-dominated sets, with the synthetic scaling study providing that evidence (Appendix~\ref{app:target_effect}). Only the four ViT checkpoints and our own models can be rescored.

More broadly, the field needs larger, more crystallographically diverse benchmarks with reproducible inclusion rules and routine release of runnable checkpoints, so published systems can be rescored rather than compared through incompatible numbers. The artifacts released here (Appendix~\ref{app:artifacts}) are offered toward that end, and toward treating evaluation as a scientific object rather than a single aggregate score.

\label{endofmain}
\begin{ack}
The authors acknowledge the Texas Advanced Computing Center (TACC) at The University of Texas at Austin (\url{http://www.tacc.utexas.edu}) for computational resources, made available to NIST under contract 1333ND25PNB180410, that contributed to the results reported here. Additional computational resources were provided through ACCESS allocation PHY250007, ``Applications of AI to Diffraction.'' Edward G. Friedman and Elizabeth J. Baggett were supported by the Center for High Resolution Neutron Scattering, a partnership between the National Institute of Standards and Technology and the National Science Foundation under Agreement No.\ DMR-2010792. 
\end{ack}

\bibliographystyle{plainnat}
\bibliography{references_revised}

\clearpage
\appendix

\section*{Appendix Contents}
\begin{itemize}[label={},leftmargin=2.6em,itemsep=1pt,parsep=0pt,topsep=2pt]
\item[{\makebox[2.2em][l]{\ref{app:mapping_build}}}] Mapping Construction and Verification \dotfill \pageref{app:mapping_build}
\item[{\makebox[2.2em][l]{\ref{app:hex_trig}}}] The Hexagonal/Trigonal Extinction-Class Pair \dotfill \pageref{app:hex_trig}
\item[{\makebox[2.2em][l]{\ref{app:target_effect}}}] Space-Group versus Extinction-Group Target \dotfill \pageref{app:target_effect}
\item[{\makebox[2.2em][l]{\ref{app:method}}}] Construction Method: Full Step Descriptions \dotfill \pageref{app:method}
\item[{\makebox[2.2em][l]{\ref{app:rruff_details}}}] Additional RRUFF Details and Discussion \dotfill \pageref{app:rruff_details}
\item[{\makebox[2.2em][l]{\ref{app:opxrd}}}] Additional opXRD Details and Discussion \dotfill \pageref{app:opxrd}
\item[{\makebox[2.2em][l]{\ref{app:dag_nulls}}}] Reference Distributions for the Hierarchy-Aware Metric \dotfill \pageref{app:dag_nulls}
\item[{\makebox[2.2em][l]{\ref{app:hier_metrics}}}] Relation to Prior Hierarchical-Classification Metrics \dotfill \pageref{app:hier_metrics}
\item[{\makebox[2.2em][l]{\ref{app:ci}}}] Bootstrap Confidence Intervals \dotfill \pageref{app:ci}
\item[{\makebox[2.2em][l]{\ref{app:cnn}}}] The 1D Residual CNN Control \dotfill \pageref{app:cnn}
\item[{\makebox[2.2em][l]{\ref{app:artifacts}}}] Released Artifacts \dotfill \pageref{app:artifacts}
\end{itemize}

\section{Mapping Construction and Verification}
\label{app:mapping_build}

\paragraph{Extinction groups in plain terms.} A screw axis or glide plane makes every reflection in some family with a particular index parity vanish, and two space groups whose symmetry elements erase exactly the same reflections leave identical fingerprints in a powder pattern; the extinction group is the name for the set of space groups sharing that fingerprint. Space groups $P2_1$ (No.~4) and $P2_1/m$ (No.~11) share a class for this reason: the screw axis produces the same absences in both, and the added mirror produces none. Extinction groups therefore define the finest evaluative label space that PXRD can justify, and a benchmark that scores in full 230-way space-group space risks penalizing a physically correct prediction simply because the measurement cannot distinguish two classes.

\paragraph{Construction.} The extinction-group mapping released with this paper is built from the extinction-symbol table
distributed with EXPO~\citep{altomare_expo}, which derives from \textit{International Tables for
Crystallography}~\citep{int_tables_a} and was first released in machine-usable form with Baggett et
al.~\citep{baggett2026attention}. The construction is deterministic: we fetch and parse that table,
expand its range and list notation into explicit space-group numbers, merge symbols carrying identical
space-group sets into canonical classes, assign stable integer labels, annotate crystal system and Laue
class, invert the result into a complete 230-row lookup, and cross-check the assignments against
\textit{International Tables}, whose treatment of extinction symbols has itself changed across editions.

The released artifact is provided as JSON and CSV. Each extinction group carries an integer label, a
canonical standard crystallographic symbol, and the associated space-group numbers; the inverted
230-row table gives, for every space group, its extinction group, Laue group, and crystal system. The
integer labels are stable and are the label space used by every benchmark and metric in this paper, so
predictions produced under the mapping remain comparable across studies that adopt it.

\section{The Hexagonal/Trigonal Extinction-Class Pair}
\label{app:hex_trig}

The merge referred to in Sections~\ref{sec:background} and~\ref{sec:dagmetric} is a taxonomy choice for the condensed subgroup graph, not an inference from model performance, and the 99-way label space retains both classes. The two classes are $P\text{-}\text{-}c$ (space groups 159, 163, 186, 190, and 194) and $P\text{-}c\text{-}$ (space groups 158, 165, 185, 188, and 193). Indexing powder lines by the hexagonal invariant $S = h^2 + hk + k^2$, the reflection conditions tabulated in \textit{International Tables}~\citep{int_tables_a} place the $P\text{-}c\text{-}$ absences on the $S = 1, 4, 9, \ldots$ family at odd $l$ and the $P\text{-}\text{-}c$ absences on the $S = 3, 12, 27, \ldots$ family at odd $l$. The two absence sets are the same pattern related by $S \rightarrow 3S$, the $30^{\circ}$ basal rotation that rescales the in-plane cell edge by $\sqrt{3}$, so the difference between the classes is a basal-direction convention rather than a different kind of absence.

The limit should be stated precisely. The distinction is not erased by powder averaging; it rests entirely on one low-angle reflection family ($10.1$ absent versus $11.1$ absent), so it is recoverable in principle yet fragile on scans where that single line is weak, overlapped, or outside the measured window. We merge the pair in the condensed graph because the two classes are additionally related by $t$-subgroup relations in both directions, which makes them incomparable in the hierarchy and requires condensation to obtain a directed acyclic graph.

The footprint on the benchmarks is exact. Extinction group 74 ($P\text{-}c\text{-}$) never occurs as a true label in RRUFF-473 or RRUFF-325, while extinction group 73 ($P\text{-}\text{-}c$) occurs 21 and 17 times. The merge therefore changes no reported accuracy; its only effect is that a prediction of group 74 against a true group 73, possible in at most 21 of 473 and 17 of 325 cases, is scored at DAG distance 0 rather than as a near miss. The explicit reflection-condition enumeration is included in the released code.

\section{Space-Group versus Extinction-Group Target}
\label{app:target_effect}

\paragraph{Experimental setup.} Table~\ref{tab:target_effect} gives the matched, convergence-controlled comparison summarized in \S\ref{sec:mapping}. Both arms were trained with the pipeline and two-million-pattern synthetic corpus of \citet{baggett2026attention} and share the same patterns, split, ViT backbone, initialization, optimizer, augmentations, training budget, and validation-selection rule; only the target and output layer differ. The direct extinction-group arm corresponds to the extinction-group model of Baggett et al.; the space-group arm was trained to convergence under the identical recipe so that the comparison is not confounded by training budget.
The space-group model is read three ways: in its own 230-way space, after its probabilities are summed into the 99-way extinction-group space, and against the model trained directly on extinction groups.

\paragraph{Scaling study and transfer rows.} The matched scaling study is the informative result. The Top-1 advantage of direct extinction-group training over the collapsed space-group model is +0.59, +1.45, and +2.77 points at one-half, one-quarter, and one-eighth training scale, respectively, against +0.86 at full scale, so direct extinction-group supervision is primarily a sample-efficiency advantage, and as the space-group corpus grows the space-group model learns much of the extinction-group equivalence structure implicitly. The transfer rows score both uncalibrated synthetic-only arms on identical RRUFF inputs in the common extinction-group space; they are not comparable with the calibrated checkpoints of Table~\ref{tab:rank_disagreement}, and every paired-bootstrap interval includes zero, so these small, domain-shift-dominated sets serve as a transfer check rather than as an estimate of the target-design effect.

\begin{table}[!htbp]
\centering
\small
\setlength{\tabcolsep}{4pt}
\caption{Matched comparison of a 230-way space-group (SG) model and a 99-way extinction-group (EG) model. ``Collapsed SG'' scores the SG model after its probabilities are summed into EG space. All accuracy cells are Top-1 / Top-3 / Top-5 (\%). Advantage is direct EG minus collapsed SG in percentage points; RRUFF advantages are Top-1 with paired-bootstrap 95\% CI and exact McNemar $p$ (Top-3 and Top-5 differences on both benchmarks likewise have $p \ge 0.42$ and intervals containing zero).}
\label{tab:target_effect}
\begin{tabularx}{\linewidth}{@{}lYYYY@{}}
\toprule
Setting & SG model, SG space & Collapsed SG, EG space & Direct EG & Direct-EG advantage \\
\midrule
Synthetic validation, full scale & 15.17 / 27.81 / 35.25 & 18.46 / 33.27 / 42.35 & 19.32 / 34.28 / 43.38 & +0.86 / +1.01 / +1.03 \\
Synthetic, $\nicefrac{1}{2}$ scale (Top-1) & --- & --- & --- & +0.59 \\
Synthetic, $\nicefrac{1}{4}$ scale (Top-1) & --- & --- & --- & +1.45 \\
Synthetic, $\nicefrac{1}{8}$ scale (Top-1) & --- & --- & --- & +2.77 \\
RRUFF-473 & --- & 3.38 / 5.92 / 8.25 & 3.59 / 5.71 / 8.46 & +0.21 [$-1.27$, 1.69], $p=1.00$ \\
RRUFF-325 & --- & 1.54 / 4.31 / 7.08 & 2.77 / 5.23 / 8.31 & +1.23 [$-0.31$, 2.77], $p=0.219$ \\
\bottomrule
\end{tabularx}
\end{table}

\section{Construction Method: Full Step Descriptions}
\label{app:method}
A source archive is a collection of heterogeneous measurements, not a benchmark: it may contain multiple scans per material, incomplete or multiphase labels, mixed wavelengths, exact duplicates, no canonical split, and no duplicate policy. The construction method turns such an archive into a fixed evaluation set whose inclusion rules are specified as code and reproduce from a frozen snapshot. Its steps are:
\begin{enumerate}[leftmargin=*,nosep]
\item \textbf{Single assignable label.} Retain records with one phase whose space group is stored/derivable and map it to the extinction-group label space of \S\ref{sec:mapping}. Multiphase records without a designated primary phase are excluded, because phase order cannot be assumed to identify the dominant diffraction signal.
\item \textbf{Wavelength and grid gate.} Select scans collected at the target wavelength (Cu-K$\alpha$, either an explicit K$\alpha_1$/K$\alpha_2$ pair or the conventional 2:1 weighted average), require complete angular support at native resolution over the evaluation window, and resample each onto a standard $2\theta$ grid.
\item \textbf{Structural grouping.} Assign each record a structural-family label, used both to group structurally near-identical records so that they are not counted as independent and to stratify results. The label source is archive-specific: a mineralogical family when the archive supplies one, otherwise a structure-derived community assignment.
\item \textbf{Family coherence.} Keep only families that are internally consistent under within-family diagnostics (full-pattern correlation, low-angle correlation, and a rank-1 reconstruction angle), separate distinct variants by label and lattice-compatible clustering, and admit singletons only when they remain similar to the family core in both lattice parameters and pattern shape.
\item \textbf{Exact-pattern deduplication}, conflict-aware: byte-identical patterns are reduced to one representative, and groups whose identical patterns carry conflicting labels are excluded.
\item \textbf{Per-pattern strata.} Attach nuisance-fit severity from classical Rietveld profile fitting under fixed $R_\mathrm{wp}$ cutoffs, and record any measurement stratum that varies within the archive (for example wavelength mode) so it can be reported separately.
\item \textbf{Freeze and audit.} Release the exact record identifiers tied to a snapshot date, the curation code with explicit thresholds, and a threshold-sensitivity sweep; if the upstream archive changes, the frozen identifier list is the reference.
\end{enumerate}
Steps 1, 2, and 7 are identical for every archive; steps 3--6 are where a source can require a different instantiation, and Figure~\ref{fig:pipelines} records, step by step, how each was instantiated on the two archives.

\section{Additional RRUFF Details and Discussion}
\label{app:rruff_details}

This appendix gives the RRUFF instantiation of the method of Appendix~\ref{app:method}, summarized in \S\ref{sec:rruff_instance} and Figure~\ref{fig:pipelines}, the released defaults, the robustness summaries that support the benchmark-design choices, and the coverage rationale.

\paragraph{Source and label assignment.} The frozen visible-filtered Cu-K$\alpha$ source pool contains 1{,}981 scans. RRUFF names a primary mineral phase for every entry, and its space group is mapped to the 99-way extinction-group label space before the coherence and quality gates are applied. The scans may contain secondary phases or impurities, but RRUFF identifies the primary mineral.

\paragraph{Curation pipeline.} The RRUFF instantiation of the method of Appendix~\ref{app:method} is as follows. The reduction to RRUFF-473 is an explicit family-coherence curation step, not a metadata screen: retained families must remain internally consistent in lattice-scale metadata and in full-pattern / low-angle correlation and reconstruction diagnostics.
\begin{enumerate}[label=(\alph*),leftmargin=*,nosep]
\item \textbf{Filter and resample.} Select RRUFF scans collected at Cu-like wavelengths ($\lambda \approx 1.54$\,\AA) and resample each onto a standard 8501-point $2\theta$ grid (5.00$^{\circ}$--90.00$^{\circ}$, 0.01$^{\circ}$ step).
\item \textbf{Group by mineral family} and use repeated families to decide which minerals support a stable real-data benchmark at all.
\item \textbf{Keep only stable families} using within-family consistency diagnostics: full-pattern Pearson correlation, low-angle correlation, and a rank-1 reconstruction angle derived from the leading singular vector of the within-family pattern matrix.
\item \textbf{Separate distinct variants} within retained families by \textit{mineral + space-group} assignment and by tight lattice-compatible clusters.
\item \textbf{Keep the coherent core algorithmically}: admit singleton clusters only when they remain similar to the main family cluster in both lattice parameters and pattern shape. This yields \textbf{RRUFF-473}.
\item \textbf{Assign nuisance-fit strata} via classical Rietveld profile fitting, i.e., full-pattern least-squares refinement of a crystallographic model against the observed diffraction trace: \textit{recoverable} (172), meaning the pattern is difficult but still reasonably explained by a standard refinement; \textit{usable} (153), meaning the fit is cleaner and more deployment-like; \textit{poor} (74), meaning substantial mismatch remains after fitting; and \textit{catastrophic} (74), meaning the classical fit fails badly enough that the pattern should be treated as a stress case rather than a routine one.
\item \textbf{Apply fixed $R_\mathrm{wp}$ cutoffs} ($R_\mathrm{wp} \le 0.12$ for the higher-quality usable-or-better subset; $0.12 < R_\mathrm{wp} \le 0.22$ for the recoverable stratum) to retain usable and recoverable patterns, producing \textbf{RRUFF-325}.
\end{enumerate}

\paragraph{Released benchmark defaults.} The curation defaults are explicit rather than hidden in code: the family-coherence stage uses fixed thresholds on lattice distance, full-pattern and low-angle correlation, anchor-cell agreement, and partial-cluster tie-break margin, and the nuisance-fit labels are deterministic $R_\mathrm{wp}$ cutoffs (usable-or-better $\le 0.12$, recoverable $\le 0.22$, poor $\le 0.50$, catastrophic above). Exact threshold values, per-scan QC fields, and the fitted lattice/profile/background parameters are listed in the released manifest. We treat those released refinement outputs as fixed benchmark metadata; the robustness analyses below therefore address threshold sensitivity of the benchmark definitions, not invariance across alternative refinement workflows. For reference, the 74 catastrophic patterns of RRUFF-473 span 13 unique extinction groups and 38 minerals, and the 172 recoverable patterns span 37 extinction groups and 111 minerals.

\paragraph{Benchmark stability.} A deterministic benchmark must be stable under reasonable variation of its thresholds. Sweeping the RRUFF-325 nuisance-fit cutoffs, the retained subset size ranges from 271 to 359 patterns while the calibrated reference model remains stable at 9.9--10.5\% Top-1 and 41.0--45.4\% Top-5---conclusions drawn on RRUFF-325 are therefore not brittle to cutoff choice. A local one-at-a-time sweep around the released family-coherence defaults left the benchmark core largely unchanged; only the loosest tested settings introduced a handful of singleton admissions or one-to-two marginal swaps (Table~\ref{tab:threshold_sensitivity}). The per-pattern severity labels are themselves a released artifact, so a third party can apply stratified analysis to their own model's predictions on RRUFF-473 without redoing any Rietveld fits.

\paragraph{Coverage and intended use.} Both benchmarks are fixed evaluation sets for extinction-group classification on Cu-K$\alpha$ laboratory PXRD, not general-purpose training corpora, and contain only experimental RRUFF measurements; the synthetic corpus used to train the case-study models is training data, not part of the release. To our knowledge, RRUFF remains the largest open experimental powder-diffraction source with broadly usable primary-phase labels, but the archive is strongly imbalanced and lacks complete extinction-group coverage, so a uniform experimental split would leave many classes with too few examples. These limitations are what motivate a fixed benchmark. Existing studies use different RRUFF subsets and preprocessing, preventing direct comparison; InformedXRD freezes the record identifiers and deterministic inclusion rules so that models can be tested on the same experimental cases. Until a larger labeled collection exists, a clearly bounded common benchmark is preferable to investigator-selected subsets, and balanced simulation can cover all 99 labels but cannot replace missing experimental data.

\begin{table}[!htbp]
\centering
\small
\setlength{\tabcolsep}{4pt}
\begin{threeparttable}
\caption{Benchmark audit against prior PXRD-ML evaluation sets. The main point is not raw score comparison but the mismatch in target taxonomy, filtering, difficult-case retention, and reproducibility.}
\label{tab:prior_studies_comp}

\begin{tabularx}{\linewidth}{@{}lYYYY@{}}
\toprule
 & Schopmans et al. & SimXRD & RRUFF-473 & RRUFF-325 \\
\midrule
Patterns Count & 942 & $\sim$3000 & 473 & 325 \\
Classification target & 145 SGs & 230 SGs & 99 EGs & 99 EGs \\
Curation method & Hand-filtered & Partially documented & Algorithmic & Algorithmic \\
Unique minerals & --- & --- & 227 & 168 \\
Unique extinction groups & --- & --- & 46 & 45 \\
Difficult cases retained?
& No (85 SGs excluded)
& Unknown
& Yes
& No\tnote{a} \\
Recoverable / Usable / Poor / Catastrophic & --- & --- & 172 / 153 / 74 / 74 & 172 / 153 / 0 / 0 \\
Reproducible from raw RRUFF? & No & No & Yes & Yes \\
\bottomrule
\end{tabularx}

\begin{tablenotes}
\footnotesize
\item[a] Restricted to the higher-quality usable and recoverable strata.
\end{tablenotes}

\end{threeparttable}
\end{table}

\begin{table}[!htbp]
\centering
\small
\setlength{\tabcolsep}{4pt}
\caption{Local one-at-a-time sensitivity of the family-coherence curation defaults. Reported variants are the only tested perturbations that changed benchmark composition beyond the baseline 471/473 core match.}
\label{tab:threshold_sensitivity}
\begin{tabularx}{\linewidth}{@{}lYYYY@{}}
\toprule
Variant & Selected & Intersection & Selected-only & Reference-only \\
\midrule
Baseline & 471 & 471 & 0 & 0 \\
Anchor full threshold $0.86$ & 473 & 471 & 2 & 0 \\
Anchor cell threshold $0.0020$ & 472 & 471 & 1 & 0 \\
Partial tie-break margin $0.04$ & 471 & 469 & 2 & 2 \\
Lattice threshold $0.0006$ & 471 & 470 & 1 & 1 \\
\bottomrule
\end{tabularx}
\end{table}

\begin{table}[!htbp]
\centering
\small
\setlength{\tabcolsep}{4pt}
\caption{Benchmark-matched prior-only controls. ``RRUFF prior'' denotes the empirical extinction-group distribution of the named benchmark itself rather than the aggregate synthetic-training prior.}
\label{tab:benchmark_prior_controls}
\begin{tabularx}{\linewidth}{@{}lYYY@{}}
\toprule
Benchmark / stratum & Top-1 & Top-5 & Top-10 \\
\midrule
RRUFF-473 prior & 12.9 & 51.4 & 68.7 \\
RRUFF-325 prior & 12.6 & 43.7 & 62.8 \\
Recoverable prior & 12.2 & 41.3 & --- \\
Usable-or-better prior & 15.7 & 48.4 & --- \\
Poor prior & 16.2 & 66.2 & --- \\
Catastrophic prior & 28.4 & 75.7 & --- \\
\bottomrule
\end{tabularx}
\end{table}

\section{Additional opXRD Details and Discussion}
\label{app:opxrd}

This appendix gives the gate table behind Figure~\ref{fig:pipelines} (right), the reasons the benchmark was not built on opXRD and the two sets are not merged, and the evaluation of the six XRDBench architectures on opXRD-98, the small example constructed in \S\ref{sec:opxrd_instance}. The exact record manifest and processing scripts implement the freeze-and-audit step.

\paragraph{Complementarity and processing.} opXRD~\citep{hollarek2025opxrd} is an open archive of 92{,}552 experimental powder patterns from multiple laboratories, still growing, and parsing cleanly enough that we built pymatgen structures~\citep{ong2013pymatgen} directly from the shipped cells and sites. opXRD and InformedXRD are complementary rather than competing---one is an experimental data archive, the other an evaluation protocol---so the comparison we can offer is not feature-by-feature but what happens when the protocol is applied to the archive. The exact list of records used and the processing scripts are released with the paper.

\paragraph{One protocol step instantiated differently.} Steps 1, 2, and 7 transfer unchanged; steps 3--6 are re-instantiated, as \S\ref{sec:construction} and Appendix~\ref{app:method} anticipate, and this is one concrete test of the design's generality. RRUFF supplies a mineralogical family, which the protocol uses both to group structurally near-identical records so that they are not counted as independent and to stratify results. opXRD carries no such label, so we projected its structures into the frozen ICSD structural map of \citet{nguyen2026icsdmap}, whose author list overlaps ours. That map embeds published ICSD structures in a similarity space and partitions them into communities of structurally related materials; a record lying inside its community's typical spread is in-basin, one at the edge is frontier, which separates well-precedented structure types from unusual ones. The two jobs the mineral family does on RRUFF are therefore done from structure alone: 95 of the 98 records carry an assignment (67 in-basin, 28 frontier), family coherence becomes distance to a community's own 95th-percentile spread, which stratifies rather than excludes, and structures outside the published projection scope are reported as explicit errors rather than silently labeled outliers. Deduplication runs as an explicit byte-identical pass rather than being absorbed into coherence, and with no Rietveld fits the wavelength mode replaces $R_\mathrm{wp}$ severity as the reported stratum. The role of each step in the protocol is unchanged; only the source of the label differs. A community assignment does not by itself confirm that a measured pattern matches its deposited structure, which remains a separate profile-consistency step. Because opXRD has no Rietveld-fit severity labels, this instantiation does not produce a second quality-filtered set; opXRD-98 remains a stress test rather than a benchmark.

\paragraph{Why the benchmark was not built on it.} Our task is one-pattern, one-label supervised classification, and that is where the archive is currently thin once filtered (Table~\ref{tab:opxrd_gates}). We chose 10--80$^{\circ}$ to match the SimXRD input range, not to maximize the count; at 5--90, 5--80, 10--90, and 15--80$^{\circ}$ the counts are 13, 39, 50, and 139. RRUFF also names a primary mineral phase for every entry; opXRD's 320 multiphase symmetry-complete records do not list phase order, and order cannot be assumed to identify the dominant phase, so they cannot enter a one-label set. Those records could support a set-valued evaluation, which we think interesting, but they are not one-label examples.

\begin{table}[!htbp]
\centering
\small
\setlength{\tabcolsep}{4pt}
\caption{Filtering gates applied to opXRD to obtain opXRD-98, a one-pattern, one-label Cu-K$\alpha$ set comparable with the RRUFF benchmarks.}
\label{tab:opxrd_gates}
\begin{tabularx}{\linewidth}{@{}Xr@{}}
\toprule
Gate & Records remaining \\
\midrule
Initial archive & 92{,}552 \\
Declares a phase & 2{,}179 \\
Space group assignable to each phase (316 records with all phase labels stored explicitly; remainder derived from structure with spglib~\citep{togo2024spglib}) & 1{,}228 \\
Single phase with lattice and derivable space group (891 from one source) & 908 \\
Cu-K$\alpha$ gate: explicit K$\alpha_1$/K$\alpha_2$ pair or the conventional 2:1 weighted average & 229 \\
RRUFF angular support at native resolution, 10--80$^{\circ}$ & 98 (28 space groups, 23 extinction groups) \\
\bottomrule
\end{tabularx}
\end{table}

\paragraph{Why the two are not merged.} The direct structural overlap is 6 opXRD records across 23 RRUFF entries, so an intersection is empty for practical purposes. A union would be worse: the label distributions differ measurably (space-group total-variation distance 0.391, Jensen--Shannon divergence 0.216 bits), and the two sets differ in phase semantics, angular support, instrument distribution, and wavelength mode, so pooling would confound source with label distribution---the failure mode this paper argues makes cross-study numbers uninterpretable. Reporting them as separate stratified sets is what our own protocol requires.

\paragraph{What opXRD-98 shows.} After exact 0.01-to-0.02$^{\circ}$ decimation and matched normalization, the six SimXRD-trained XRDBench architectures reach extinction-group Top-1 of 4.08--10.20\% and Top-5 of 22.45--38.78\% on opXRD-98 (Table~\ref{tab:opxrd_models}). We present this as a stress test, not a benchmark, and report its limits: an input-blind predictor using opXRD-98's own label marginal scores 22.45/53.06/66.33\% Top-1/3/5, above every model, and no model improves on its matched DAG null (2.820--2.840; the exact uniform null is 2.876 with 32.62\% within two hops).

\begin{table}[!htbp]
\centering
\small
\setlength{\tabcolsep}{4pt}
\caption{Six XRDBench architectures on opXRD-98 (98 Cu-K$\alpha$ records), rescored in extinction-group space. Hierarchy columns are over wrong Top-1 predictions.}
\label{tab:opxrd_models}
\begin{tabularx}{0.9\linewidth}{@{}lYYY@{}}
\toprule
Model & EG Top-1 / 3 / 5 (\%) & $\le$2 hops (\%) & Mean DAG dist. \\
\midrule
PatchTST    & \textbf{10.20} / 20.41 / 34.69 & \textbf{32.95} & \textbf{3.034} \\
CNN11       & 6.12 / 20.41 / 26.53 & 31.52 & 3.174 \\
CNN10       & 5.10 / 19.39 / \textbf{38.78} & 29.03 & 3.258 \\
BiLSTM      & 5.10 / 15.31 / 22.45 & 23.66 & 3.441 \\
BiGRU       & 4.08 / 17.35 / 24.49 & 20.21 & 3.394 \\
Transformer & 4.08 / \textbf{22.45} / 31.63 & 17.02 & 3.745 \\
\midrule
Input-blind (opXRD-98 marginal) & 22.45 / 53.06 / 66.33 & --- & --- \\
\bottomrule
\end{tabularx}
\end{table}

Top-5 is also far higher on the 33 weighted-average records (39.39--51.52\%) than on the 65 explicit-pair records (13.85--33.85\%); wavelength mode is confounded with source and label composition here, so this is descriptive rather than causal, but it shows that wavelength must remain a reported stratum. Finally, one finding we offer back to the archive: among the 908 single-phase records, 13 groups of byte-identical raw patterns (27 records) carry 6 conflicting space-group/extinction-group labels, leaving a conflict-aware exact-pattern set of 888 records.

\section{Reference Distributions for the Hierarchy-Aware Metric}
\label{app:dag_nulls}

Table~\ref{tab:dag_nulls} lists the four reference distributions defined in \S\ref{sec:dagmetric}. The uniform null is model-independent and exact. It uses the 98 wrong labels rather than all 99 so that it is comparable with the observed statistic, which is averaged over the model's wrong predictions only; letting the null draw the correct label would hand it a free distance of zero. The model-matched null is recomputed over each model's own error set and is the reference against which the observed summaries in Tables~\ref{tab:rank_disagreement} and~\ref{tab:xrdbench} should be read. The frequency-matched null draws the wrong label from the benchmark's own extinction-group marginal. The most-frequent-group predictor is included to show why a low mean distance is not by itself evidence of crystallographically informed errors.

\begin{table}[!htbp]
\centering
\small
\setlength{\tabcolsep}{4pt}
\caption{Reference distributions for the hierarchy-aware metric and three RRUFF-325 case-study models. Direction is the descendant / ancestor / branch-jump split over wrong predictions.}
\label{tab:dag_nulls}
\begin{tabularx}{\linewidth}{@{}lYYY@{}}
\toprule
Reference or model & Mean DAG distance & Within 2 hops (\%) & Descendant / ancestor / branch (\%) \\
\midrule
Uniform null, RRUFF-473 & 2.997 & 29.76 & 8.65 / 35.61 / 55.69 \\
Uniform null, RRUFF-325 & 2.944 & 30.65 & 10.04 / 29.53 / 60.38 \\
Frequency-matched null, RRUFF-473 & 2.692 & 42.15 & --- \\
Frequency-matched null, RRUFF-325 & 2.656 & 43.35 & --- \\
Most-frequent-group predictor, RRUFF-325 & 2.116 & 63.4 & --- \\
\midrule
Final mixed-data ViT, RRUFF-325 & 2.46 & 51.55 & 56.7 / 10.7 / 32.6 \\
\quad its model-matched null & 2.981 & 28.92 & 10.8 / 27.4 / 61.8 \\
Preferred-orientation-only ViT, RRUFF-325 & 2.51 & 53.0 & --- / --- / 34.1 \\
\quad its model-matched null & 2.973 & 28.85 & 11.1 / 25.9 / 62.9 \\
1D residual CNN (epoch 6), RRUFF-325 & 2.921 & 28.05 & 27.4 / 18.8 / 53.8 \\
\quad its model-matched null & 2.937 & 30.45 & 10.5 / 27.5 / 61.9 \\
\bottomrule
\end{tabularx}
\end{table}
\noindent Model-matched nulls are reported for the checkpoints whose per-case predictions are in the released failure-mode artifacts; the uniform null is exact over $473 \times 98 = 46{,}354$ and $325 \times 98 = 31{,}850$ wrong-label pairs.

\begin{figure}[!htbp]
\centering
\includegraphics[width=0.85\linewidth]{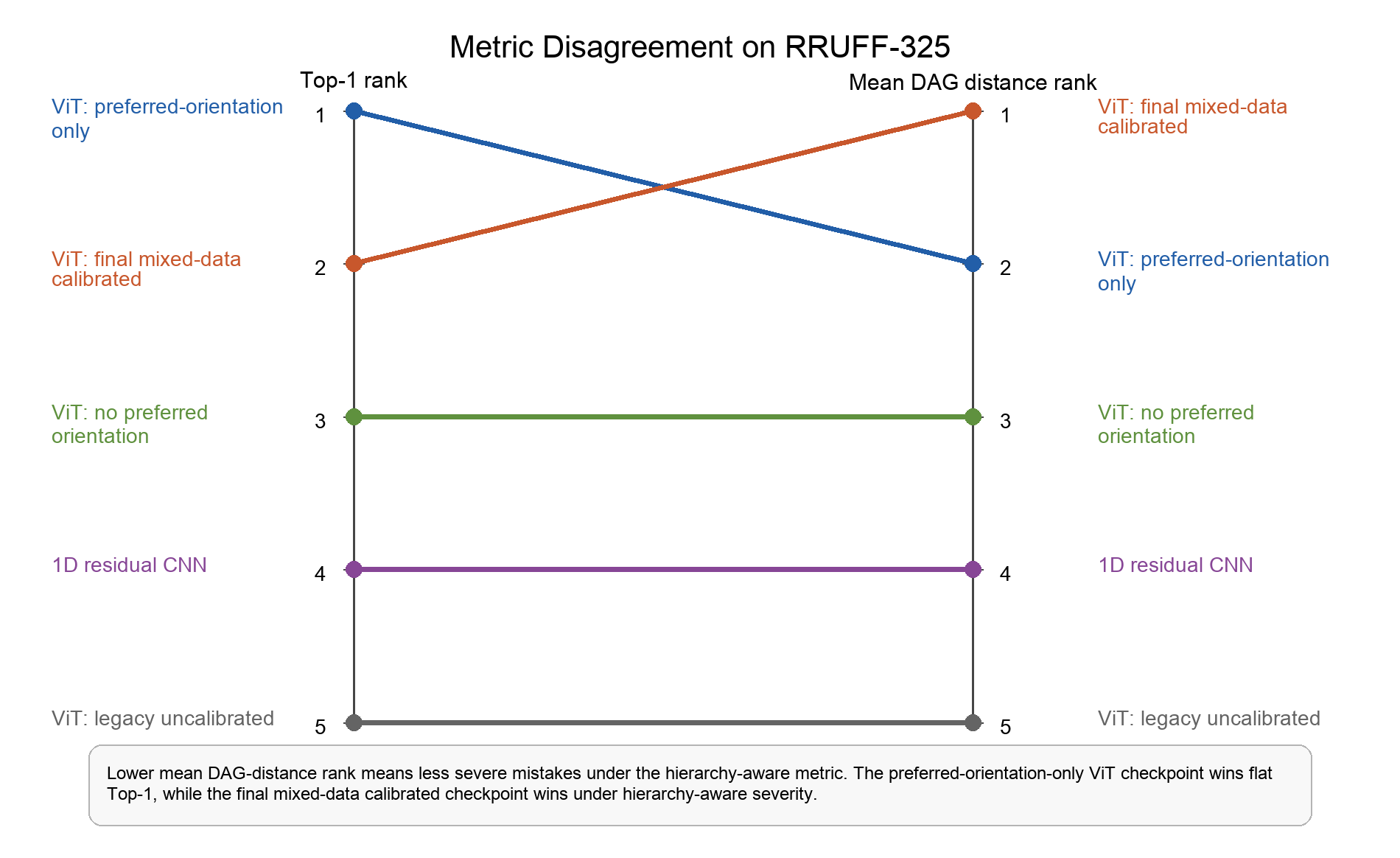}
\caption{Flat accuracy and hierarchy-aware severity rank the same five models differently. Each line is one model; its left endpoint is its Top-1 rank, and its right endpoint is its mean-DAG-distance rank on the condensed $t$-subgroup graph. Lower mean DAG distance is better. The preferred-orientation-only ViT checkpoint wins under flat Top-1, while the final mixed-data calibrated ViT checkpoint has the best raw mean-DAG-distance rank. The 1D CNN baseline remains worse on both metrics and, more importantly for this paper, demonstrates that the hierarchy-aware metric distinguishes a separate branch-jump-heavy failure regime that flat Top-1 alone would compress into a single score.}
\label{fig:metric_disagreement}
\end{figure}

\section{Relation to Prior Hierarchical-Classification Metrics}
\label{app:hier_metrics}

Hierarchical classification metrics fall into four broad families: set-overlap, distance-based, path/level-aware, and cost-weighted~\citep{silla2011survey}. We opted for a distance-based metric to answer the crystallographic question of how far the predicted symmetry is from the true one, but ancestor/overlap, level, and cost-weighted metrics are fair choices. The contribution here is the crystallographic instantiation: the condensed $t$-subgroup graph over extinction groups (a $t$-only restriction avoids conflating point-group symmetry loss with lattice-volume change, which a mixed $t/k$ graph would homogenize), the unweighted-edge choice (auditable, integer-valued, free of subjective severity weights), and the descendant/ancestor/branch-jump disaggregation that reads directly against the diffraction error model. Many scientific classification problems share this structure (biological taxonomy, medical ontologies, material hierarchies), and we hope this work serves as an example for other domains where flat accuracy remains the default.

\section{Bootstrap Confidence Intervals}
\label{app:ci}
Table~\ref{tab:ci} gives nonparametric bootstrap 95\% intervals (2{,}000 resamples over benchmark cases, seed 124) for the final mixed-data calibrated ViT checkpoint. The intervals are wide relative to the between-checkpoint differences in Table~\ref{tab:rank_disagreement}, which is why the paper relies on paired tests on the same cases rather than on overlapping marginal intervals.

\begin{table}[!htbp]
\centering
\small
\caption{Bootstrap 95\% confidence intervals for the final mixed-data calibrated ViT checkpoint (point estimate [CI]).}
\label{tab:ci}
\begin{tabularx}{0.8\linewidth}{@{}lYY@{}}
\toprule
Benchmark & Top-1 (\%) & Top-5 (\%) \\
\midrule
RRUFF-473 ($n=473$) & 9.94 [7.61, 12.68] & 50.74 [46.29, 55.18] \\
RRUFF-325 ($n=325$) & 10.46 [7.38, 13.85] & 43.69 [38.46, 48.92] \\
\bottomrule
\end{tabularx}
\end{table}

\section{The 1D Residual CNN Control}
\label{app:cnn}

The control is a 1D residual CNN with 646{,}596 parameters: a single-channel stem convolution (kernel 13, stride 1) with LayerNorm, GELU, and dropout 0.05; max-pooling with kernel and stride 7; four residual stages of two blocks each with channels $[32, 64, 64, 64]$, kernels $[5, 7, 17, 13]$, and strides $[4, 4, 5, 3]$, each block using two convolutions with LayerNorm and dropout, a projected residual where the shape changes, and GELU; adaptive max-pooling; and a linear 99-class extinction-group head. It was trained with AdamW (learning rate $10^{-3}$, weight decay $10^{-4}$), batch size 128, cross-entropy loss, and reduce-on-plateau learning-rate decay (factor 0.3, patience 2, floor $10^{-6}$), seed 1337, on a 2.35M-pattern RRUFF-conditioned synthetic corpus (1{,}878{,}370 training and 237{,}056 validation patterns) on the same 8501-point 5--90$^{\circ}$ grid as the benchmarks. Its exact epoch-6 checkpoint, with weights and configuration metadata, is included in the released model artifact, so the evaluated model is recoverable without the ViT code path. Table~\ref{tab:cnn_convergence} reports the convergence test of \S\ref{sec:disagree}: the submitted lineage was resumed for 40 further epochs with checkpoint selection on held-out synthetic validation loss alone, and the validation-selected epoch~46 is better on synthetic validation and worse on both real benchmarks.

\begin{table}[!htbp]
\centering
\small
\setlength{\tabcolsep}{4pt}
\caption{Convergence test for the 1D residual CNN control. Validation loss and Top-1/3/5 are on held-out synthetic data; benchmark accuracy columns are Top-1 / Top-3 / Top-5 (\%); hierarchy columns are the within-2-hop fraction and mean DAG distance over wrong predictions. Epoch 46 is the validation-loss minimum after repeated learning-rate reductions; the formal early-stopping rule (patience 8) had not fired.}
\label{tab:cnn_convergence}
\scriptsize
\begin{tabularx}{\linewidth}{@{}lYYYYYY@{}}
\toprule
Checkpoint & Val.\ loss & Val.\ Top-1/3/5 (\%) & RRUFF-473 acc. & RRUFF-473 $\le$2 hops / mean dist. & RRUFF-325 acc. & RRUFF-325 $\le$2 hops / mean dist. \\
\midrule
Submitted epoch 6 & 1.6631 & 50.25 / 75.47 / 84.85 & 9.09 / 19.66 / 28.54 & 28.84\% / 2.956 & 6.77 / 18.77 / 25.23 & 28.05\% / 2.921 \\
Validation-selected epoch 46 & 1.3948 & 58.00 / 81.30 / 89.03 & 3.59 / 8.46 / 15.64 & 17.98\% / 3.342 & 2.15 / 4.92 / 12.92 & 20.13\% / 3.255 \\
\bottomrule
\end{tabularx}
\end{table}

\section{Released Artifacts}
\label{app:artifacts}

\paragraph{Relationship with Prior Release.} Baggett et al.~\citep{baggett2026attention} first introduced and publicly released the extinction-group mapping, RRUFF-473 and RRUFF-325 with their nuisance-fit labels and curation logic, the condensed subgroup graph, and the four ViT checkpoints. This paper reuses those released components and credits them where each is introduced (\S\ref{sec:mapping}, \S\ref{sec:benchmarks}, and \S\ref{sec:dagmetric}). The new contribution here is the standardized evaluation package---the frozen artifacts, the deployable and oracle prior controls, the permutation test, the null reference distributions for the hierarchy metric, the stratified protocol, and the cross-architecture and cross-archive case studies---and the checkpoints are evidence for how that package changes what claims are justified from the same predictions. The nuisance/entropy observation and the subgroup graph appeared first in Baggett et al.; here they are load-bearing parts of a method rather than findings. opXRD-98, its record list, and its processing scripts exist only in this paper.

\paragraph{XRDBench training contract.} The six XRDBench architectures of Table~\ref{tab:xrdbench} were trained on the unchanged public SimXRD \texttt{train\_1} contract~\citep{simxrd2025}: 597{,}845 patterns, 230 space-group targets, 10--80$^{\circ}$ at 0.02$^{\circ}$, and 3{,}501 points per pattern.

\begin{figure}[!htbp]
\centering
\includegraphics[width=\linewidth]{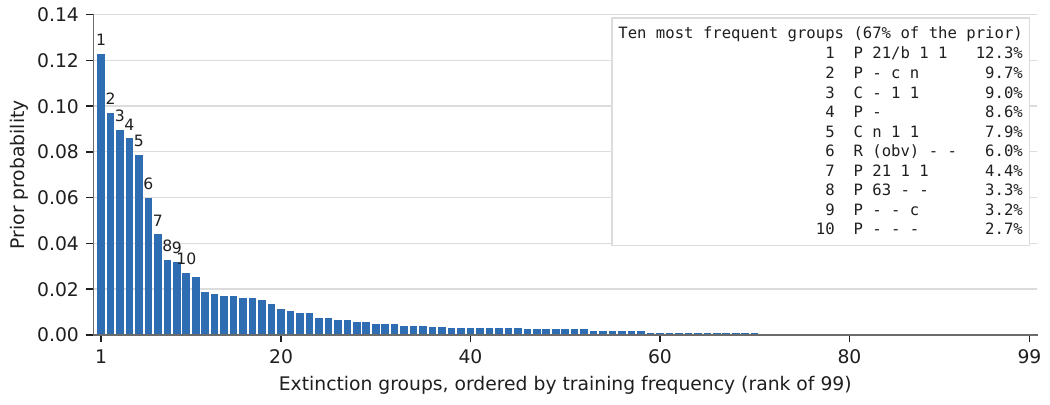}
\caption{Extinction-group imbalance in the ICSD-derived source distribution underlying the aggregate synthetic training prior used for calibration. Horizontal axis: all 99 extinction groups, ordered by training frequency; vertical axis: prior probability; the ten most frequent groups are numbered on the bars and named in the inset. Heavy class imbalance means a naive prior-frequency baseline can attain non-trivial Top-$k$ accuracy with no reference to the input.}
\label{fig:icsd_distribution}
\end{figure}

\paragraph{The aggregate prior.} We use a compact aggregate prior estimated from the synthetic training corpus used by the case-study calibration pipeline. The pattern-level corpus is not redistributed, but the released prior is the complete aggregate quantity the evaluation consumes---a 99-class prior with probabilities, log-priors, seen-class indicators, and the smoothing parameter---so every prior-only and prior-corrected number reported here is reproducible without it. Figure~\ref{fig:icsd_distribution} shows the extinction-group imbalance of the source distribution behind it.

All artifacts (mapping tables, DAG adjacency lists, stratification labels, evaluation scripts, benchmark payloads, case-study checkpoints, and the processing scripts and exact record list for the opXRD case study of Appendix~\ref{app:opxrd}) are public: the GitHub reproducibility repository at \url{https://github.com/scattering/InformedXRD} carries the curation and evaluation code, and the companion Hugging Face dataset \url{https://huggingface.co/datasets/informedxrd/informedxrd-benchmarks} and model \url{https://huggingface.co/informedxrd/informedxrd-checkpoints} repositories carry the external benchmark payloads and checkpoints. The dataset repository is released under CC-BY-4.0 and the checkpoint and code repositories under the MIT license; the underlying RRUFF and opXRD data remain under their original terms of use. The repository is deliberately self-contained: the model implementation, extinction-group lookups, benchmark builders, subgroup-graph code, ViT evaluation code, and the four ViT checkpoints are reused or adapted from the public Baggett et al.\ release, and each such file is marked with its provenance. The files specific to this paper are the evaluation documentation and manifests, dataset and model metadata, benchmark packaging, the CNN control with its standalone implementation and loader, the prior-only, permutation-test, and null-distribution scripts, the XRDBench rescoring scripts, the opXRD processing pipeline, and a worked example that takes a model's predicted probabilities, rescores a 230-way output into the 99-way extinction-group space, and reproduces a reported row end to end.

\clearpage
\section*{NeurIPS Paper Checklist}

\begin{enumerate}

\item {\bf Claims}
    \item[] Question: Do the main claims made in the abstract and introduction accurately reflect the paper's contributions and scope?
    \item[] Answer: \answerYes{}
    \item[] Justification: The abstract and the contribution list in Section~1 name the three parts of the framework (the extinction-group mapping; the benchmark-construction method with its RRUFF-473, RRUFF-325, and opXRD-98 instances; and the evaluation protocol of prior-only baselines, a label-permutation test, the hierarchy-aware metric with null references, and stratified reporting) and the demonstration on four ViT checkpoints, a CNN control, and six XRDBench architectures (Sections~\ref{sec:mapping}--\ref{sec:revealed}). Both state which components originate in Baggett et al.\ and which are new here; the ``Relationship to Baggett et al.'' paragraph and Appendix~\ref{app:artifacts} identify reused versus new components file by file. Section~1 states the scope as evaluation infrastructure rather than a new architecture.

\item {\bf Limitations}
    \item[] Question: Does the paper discuss the limitations of the work performed by the authors?
    \item[] Answer: \answerYes{}
    \item[] Justification: A dedicated ``Limitations'' paragraph in Section~\ref{sec:conclusion} collects them: coverage of only 46 and 45 of the 99 extinction groups and restriction to Cu-K$\alpha$ laboratory data and RRUFF's primary-phase labels; the inability of paired Top-$k$ tests to resolve the leading XRDBench architectures at these benchmark sizes; the final calibrated model matching or falling below the oracle prior-only control in aggregate; the unweighted, $t$-only DAG and the suggestive status of cross-architecture distance comparisons; benchmark stability checked by threshold sweeps rather than alternative refinement workflows; the transfer check of the target-choice effect resting on small, domain-shift-dominated sets; and the fact that only the four ViT checkpoints and our own models can be rescored. Section~\ref{sec:opxrd_instance} and Appendix~\ref{app:opxrd} state that opXRD-98 is a stress test rather than a benchmark.

\item {\bf Theory assumptions and proofs}
    \item[] Question: For each theoretical result, does the paper provide the full set of assumptions and a complete (and correct) proof?
    \item[] Answer: \answerNA{}
    \item[] Justification: The paper's contribution is empirical and methodological (evaluation infrastructure, benchmark construction, hierarchy-aware error analysis). It does not state or prove theorems.

\item {\bf Experimental result reproducibility}
    \item[] Question: Does the paper fully disclose all the information needed to reproduce the main experimental results of the paper to the extent that it affects the main claims and/or conclusions of the paper (regardless of whether the code and data are provided or not)?
    \item[] Answer: \answerYes{}
    \item[] Justification: The paper releases the exact RRUFF scan-ID lists for both benchmarks tied to a frozen upstream snapshot, the curation code with its explicit thresholds and $R_\mathrm{wp}$ cutoffs, the extinction-group mapping table as a JSON/CSV artifact, the condensed $t$-subgroup DAG adjacency lists, the stratification labels, the aggregate prior, the opXRD-98 record manifest and processing scripts, and the case-study checkpoints. Section~\ref{sec:rruff_instance} (``Reproducibility, coverage, and intended use'') states that the released scan-ID lists, snapshot date, curation code, and thresholds reproduce both benchmarks exactly; Appendix~\ref{app:method} gives the full step descriptions, Appendix~\ref{app:rruff_details} the released defaults and threshold sweeps, Appendix~\ref{app:opxrd} the opXRD-98 gate table, and Appendices~\ref{app:dag_nulls}, \ref{app:cnn}, and~\ref{app:artifacts} the null distributions, the CNN configuration, and the XRDBench training contract.

\item {\bf Open access to data and code}
    \item[] Question: Does the paper provide open access to the data and code, with sufficient instructions to faithfully reproduce the main experimental results, as described in supplemental material?
    \item[] Answer: \answerYes{}
    \item[] Justification: All artifacts are public. The GitHub repository (\url{https://github.com/scattering/InformedXRD}) provides the curation and evaluation code, environment files, manifests, notebooks, and a worked example that rescores a 230-way output into extinction-group space and reproduces a reported row end to end. The Hugging Face dataset repository (\url{https://huggingface.co/datasets/informedxrd/informedxrd-benchmarks}, CC-BY-4.0, with a Croissant file) and model repository (\url{https://huggingface.co/informedxrd/informedxrd-checkpoints}, MIT) carry the benchmark payloads, stratification labels, the aggregate prior, and the case-study checkpoints. Appendix~\ref{app:artifacts} lists the inventory and marks which files are reused from Baggett et al.

\item {\bf Experimental setting/details}
    \item[] Question: Does the paper specify all the training and test details (e.g., data splits, hyperparameters, how they were chosen, type of optimizer) necessary to understand the results?
    \item[] Answer: \answerYes{}
    \item[] Justification: The paper specifies the evaluation pipeline in detail: $2\theta$ resampling onto an 8501-point grid (5.00$^\circ$--90.00$^\circ$, 0.01$^\circ$ step), the Cu-K$\alpha$ wavelength filter, family-coherence thresholds and tie-break margins, the four nuisance-fit strata defined by deterministic $R_\mathrm{wp}$ cutoffs, the prior-only baseline construction, the label-permutation test, and the condensed $t$-subgroup DAG with its four reference distributions. The case-study models are the released ViT checkpoints of Baggett et al., an in-house 1D residual CNN whose architecture, optimizer, schedule, seed, and training corpus are given in Appendix~\ref{app:cnn} together with its 40-epoch convergence continuation, six architectures trained under the published XRDBench recipe on the unchanged SimXRD \texttt{train\_1} contract (Appendix~\ref{app:artifacts}), and a matched space-group/extinction-group ViT pair with a scaling study (Appendix~\ref{app:target_effect}). Upstream training procedures are documented in the cited references and the released code.

\item {\bf Experiment statistical significance}
    \item[] Question: Does the paper report error bars suitably and correctly defined or other appropriate information about the statistical significance of the experiments?
    \item[] Answer: \answerYes{}
    \item[] Justification: Model comparisons are reported with exact McNemar tests on Top-1 and Top-5 with explicit $p$-values and paired permutation tests on the hierarchy-aware summaries (Section~\ref{sec:disagree}); exact McNemar tests with Holm adjustment and 10{,}000-replicate paired-bootstrap intervals for the XRDBench architectures (Section~\ref{sec:xrdbench}); a 10{,}000-permutation label-permutation test with one-sided $p$-values (Section~\ref{sec:evidence}); bootstrap confidence intervals for the final calibrated checkpoint (Appendix~\ref{app:ci}); and exact null reference distributions for the hierarchy-aware metric (Appendix~\ref{app:dag_nulls}). Threshold-sensitivity ranges are reported as min--max bands over the swept variants, and aggregates are reported alongside their stratified decomposition.

\item {\bf Experiments compute resources}
    \item[] Question: For each experiment, does the paper provide sufficient information on the computer resources (type of compute workers, memory, time of execution) needed to reproduce the experiments?
    \item[] Answer: \answerNo{}
    \item[] Justification: Compute is reported only in part. Rescoring a released checkpoint on a benchmark takes under a minute per model on one GPU (the released XRDBench evaluation job ran on a TACC Vista GPU node), and the CNN control of Appendix~\ref{app:cnn} trains at about 15 minutes per epoch on one GPU according to the released convergence log, so the submitted epoch-6 checkpoint and the 40-epoch continuation took about 1.5 and 10 GPU-hours. The six XRDBench trainings and the matched space-group/extinction-group ViT pair with its scaling study ran on the TACC and ACCESS allocations named in the acknowledgments, but their GPU-hours were not logged and are not itemized in the paper. The upstream synthetic corpora and the Baggett et al.\ ViT checkpoints are described in the cited prior work rather than reproduced here.

\item {\bf Code of ethics}
    \item[] Question: Does the research conducted in the paper conform, in every respect, with the NeurIPS Code of Ethics \url{https://neurips.cc/public/EthicsGuidelines}?
    \item[] Answer: \answerYes{}
    \item[] Justification: The authors have read the NeurIPS Code of Ethics and the work conforms in every respect. The paper involves no human subjects, no scraped or personally identifying data, and no dual-use models. The released artifacts are derived from publicly available scientific archives (RRUFF, opXRD) and from previously released model checkpoints credited to their authors.

\item {\bf Broader impacts}
    \item[] Question: Does the paper discuss both potential positive societal impacts and negative societal impacts of the work performed?
    \item[] Answer: \answerYes{}
    \item[] Justification: The conclusion discusses the positive impact: more reliable, comparable, and physically meaningful evaluation in PXRD-ML, and the field-wide benefit of routine release of runnable checkpoints so that published systems can be rescored; Appendix~\ref{app:hier_metrics} notes that hierarchy-aware evaluation transfers to other structured-label scientific domains. The paper does not identify a negative societal impact: the artifacts are evaluation infrastructure for crystallographic classification rather than generative or decision-making systems, and the foreseeable misuse, overfitting to the released subsets, is addressed by the recommendation to report both benchmarks with stratified analysis (Section~\ref{sec:rruff_instance}).

\item {\bf Safeguards}
    \item[] Question: Does the paper describe safeguards that have been put in place for responsible release of data or models that have a high risk for misuse (e.g., pre-trained language models, image generators, or scraped datasets)?
    \item[] Answer: \answerNA{}
    \item[] Justification: The released artifacts are crystallographic evaluation infrastructure: powder-diffraction patterns of natural minerals, mapping tables, evaluation scripts, and small classifier checkpoints whose only outputs are extinction-group labels. They pose none of the misuse risks the question targets (no generative content, no scraped personal data, no dual-use capability).

\item {\bf Licenses for existing assets}
    \item[] Question: Are the creators or original owners of assets (e.g., code, data, models), used in the paper, properly credited and are the license and terms of use explicitly mentioned and properly respected?
    \item[] Answer: \answerYes{}
    \item[] Justification: The paper cites the creators of every external asset it builds on: the RRUFF mineral database, the EXPO extinction-symbol table and the International Tables for Crystallography volumes underlying the mapping and the subgroup graph, the opXRD archive, pymatgen and spglib, the SimXRD-4M simulator and XRDBench recipe, the ICSD structural map used for opXRD-98, and the Baggett et al.\ checkpoints, mapping, benchmarks, and subgroup graph. Appendix~\ref{app:artifacts} states the licenses of the released repositories (CC-BY-4.0 for the dataset, MIT for the checkpoints and the code) and that the underlying RRUFF and opXRD data remain under their original terms of use; the aggregate synthetic-training prior is released only as class probabilities, with no upstream structures redistributed.

\item {\bf New assets}
    \item[] Question: Are new assets introduced in the paper well documented and is the documentation provided alongside the assets?
    \item[] Answer: \answerYes{}
    \item[] Justification: The released benchmarks, mapping tables, DAG adjacency lists, stratification labels, aggregate prior, evaluation scripts, opXRD-98 record manifest, and case-study checkpoints are documented through (i) the dataset card in the Hugging Face dataset repository (intended use, licensing, contents) and its Croissant file with core and Responsible AI fields; (ii) the model card in the Hugging Face model repository (intended use, provenance, licensing); and (iii) the code repository's README, manifests, notebooks, per-file provenance markers distinguishing files reused from Baggett et al.\ from files new to this paper, a standalone CNN implementation and loader, and a worked example that reproduces a reported row. The documentation covers the released benchmark and evaluation assets, not every upstream training corpus used by the case-study models.

\item {\bf Crowdsourcing and research with human subjects}
    \item[] Question: For crowdsourcing experiments and research with human subjects, does the paper include the full text of instructions given to participants and screenshots, if applicable, as well as details about compensation (if any)?
    \item[] Answer: \answerNA{}
    \item[] Justification: The paper does not involve crowdsourcing or research with human subjects.

\item {\bf Institutional review board (IRB) approvals or equivalent for research with human subjects}
    \item[] Question: Does the paper describe potential risks incurred by study participants, whether such risks were disclosed to the subjects, and whether Institutional Review Board (IRB) approvals (or an equivalent approval/review based on the requirements of your country or institution) were obtained?
    \item[] Answer: \answerNA{}
    \item[] Justification: The paper does not involve crowdsourcing or research with human subjects.

\item {\bf Declaration of LLM usage}
    \item[] Question: Does the paper describe the usage of LLMs if it is an important, original, or non-standard component of the core methods in this research? Note that if the LLM is used only for writing, editing, or formatting purposes and does \emph{not} impact the core methodology, scientific rigor, or originality of the research, declaration is not required.
    \item[] Answer: \answerNA{}
    \item[] Justification: The core methodology (extinction-group mapping, benchmark construction, prior-only baselines, label-permutation test, hierarchy-aware DAG metric, stratified evaluation, and classifier evaluation) does not involve LLMs as an important, original, or non-standard component.

\end{enumerate}

\end{document}